\pdfoutput=1

\documentclass[11pt]{article}

\usepackage[final]{acl}

\usepackage{times}
\usepackage{latexsym}

\usepackage[T1]{fontenc}

\usepackage[utf8]{inputenc}

\usepackage{microtype}

\usepackage{inconsolata}

\usepackage{graphicx}

\usepackage{amsmath}
\usepackage{amssymb}
\usepackage{fdsymbol}
\usepackage{mathtools}
\usepackage{amsthm}
\usepackage{bm}
\usepackage{xfrac}

\usepackage[ruled,vlined,linesnumbered]{algorithm2e}

\usepackage{subfigure}
\usepackage{booktabs}
\usepackage{tabularx}
\usepackage{rotating}

\usepackage{xspace}
\usepackage{multirow}
\usepackage{multicol}
\usepackage{comment}
\usepackage{enumitem}
\usepackage[dvipsnames]{xcolor}
\usepackage[most]{tcolorbox}
\usepackage{pifont}

\usepackage{fontawesome5}
\usepackage{colortbl}
\colorlet{Yes}{blue!15}
\colorlet{No}{red!10}
\colorlet{Noo}{red!30}
\colorlet{Nooo}{red!40}

\newcommand{\Name}{LLMAP\xspace}
\newcommand{\AlgName}{MSGS\xspace}

\definecolor{Map}{HTML}{FFCCCC}
\definecolor{Performance}{HTML}{CCFFCC}
\definecolor{Constraint}{HTML}{CCFFFF}
\colorlet{key}{red!80!black}

\newtcolorbox{Quote}[1][]{breakable, colback=gray!10, colframe=gray!50, boxrule=0pt, leftrule=3pt, arc=0pt, outer arc=0pt, left=5pt, right=5pt, top=5pt, bottom=5pt, #1}

\usepackage{listings}
\colorlet{punct}{red!60!black}
\colorlet{numb}{magenta!80!black}

\lstdefinelanguage{json}{
    basicstyle=\small\ttfamily,
    breaklines=true,
    frame=single,
    backgroundcolor=\color{white},
    literate=
     *{0}{{{\color{numb}0}}}{1}
      {1}{{{\color{numb}1}}}{1}
      {2}{{{\color{numb}2}}}{1}
      {3}{{{\color{numb}3}}}{1}
      {4}{{{\color{numb}4}}}{1}
      {5}{{{\color{numb}5}}}{1}
      {6}{{{\color{numb}6}}}{1}
      {7}{{{\color{numb}7}}}{1}
      {8}{{{\color{numb}8}}}{1}
      {9}{{{\color{numb}9}}}{1}
      {:}{{{\color{punct}{:}}}}{1}
      {,}{{{\color{punct}{,}}}}{1}
      {\{}{{{\color{RoyalBlue}{\{}}}}{1}
      {\}}{{{\color{RoyalBlue}{\}}}}}{1}
      {[}{{{\color{RoyalBlue}{[}}}}{1}
      {]}{{{\color{RoyalBlue}{]}}}}{1},
}

\lstdefinelanguage{system prompt}{
    basicstyle=\ttfamily\small,
    breaklines=true,
    frame=single,
    backgroundcolor=\color{white},
}

\lstdefinelanguage{user prompt}{
    basicstyle=\ttfamily\small,
    breaklines=true,
    frame=single,
    backgroundcolor=\color{white},
    moredelim=[s][\color{RoyalBlue}]{\{}{\}},
}

\title{LLMAP: LLM-Assisted Multi-Objective Route Planning \\ with User Preferences}

\author{
  Liangqi Yuan$^\spadesuit$\thanks{Corresponding author.}, Dong-Jun Han$^\heartsuit$, Christopher G. Brinton$^\spadesuit$, Sabine Brunswicker$^\diamondsuit$\\
  $^\spadesuit$ School of Electrical and Computer Engineering, Purdue University, West Lafayette, USA,\\
  $^\heartsuit$ Department of Computer Science and Engineering, Yonsei University, Seoul, South Korea\\
  $^\diamondsuit$ Polytechnic Institute, Purdue University, West Lafayette, USA\\
\texttt{\{liangqiy,cgb,sbrunswi\}@purdue.edu} \\
\texttt{djh@yonsei.ac.kr} \\
}

\begin{document}
\maketitle
\begin{abstract}

The rise of large language models (LLMs) has made natural language-driven route planning an emerging research area that encompasses rich user objectives. Current research exhibits two distinct approaches: direct route planning using LLM-as-Agent and graph-based searching strategies. However, LLMs in the former approach struggle to handle extensive map data, while the latter shows limited capability in understanding natural language preferences. Additionally, a more critical challenge arises from the highly heterogeneous and unpredictable spatio-temporal distribution of users across the globe. In this paper, we introduce a novel LLM-Assisted route Planning (\Name) system that employs an LLM-as-Parser to comprehend natural language, identify tasks, and extract user preferences and recognize task dependencies, coupled with a Multi-Step Graph construction with iterative Search (\AlgName) algorithm as the underlying solver for optimal route finding. Our multi-objective optimization approach adaptively tunes objective weights to maximize points of interest (POI) quality and task completion rate while minimizing route distance, subject to three key constraints: user time limits, POI opening hours, and task dependencies. We conduct extensive experiments using 1,000 routing prompts sampled with varying complexity across 14 countries and 27 cities worldwide. The results demonstrate that our approach achieves superior performance with guarantees across multiple constraints. \footnote{Code and data are available at \url{https://github.com/liangqiyuan/LLMAP}.}

\end{abstract}

\section{Introduction}

The advancement in the natural language understanding capabilities of large language models (LLMs) has transformed many tasks from manual planning to automated understanding, reasoning, and decision making based on human natural language input \cite{song2023llm, huang2024understanding, kambhampati2024llms, yuan2025local}. For example, a transformative domain is autonomous driving \cite{sharan2023llm, zeng2024perceive}, where, for safety considerations, drivers often rely on voice control for tasks such as music playback, message responses, and navigation \cite{feng2024citybench}. Beyond predetermined itineraries, a more common scenario involves users knowing only their daily tasks, which consist of a series of points of interest (POI) types such as grocery stores, pharmacies, and banks, without specific ordering or POI selection. This situation introduces multiple trade-offs, including the balance between POI quality and distance, while considering various constraints such as user time limits and POI opening hours.

\begin{figure}[t]
    \centering
    \includegraphics[width=1\linewidth]{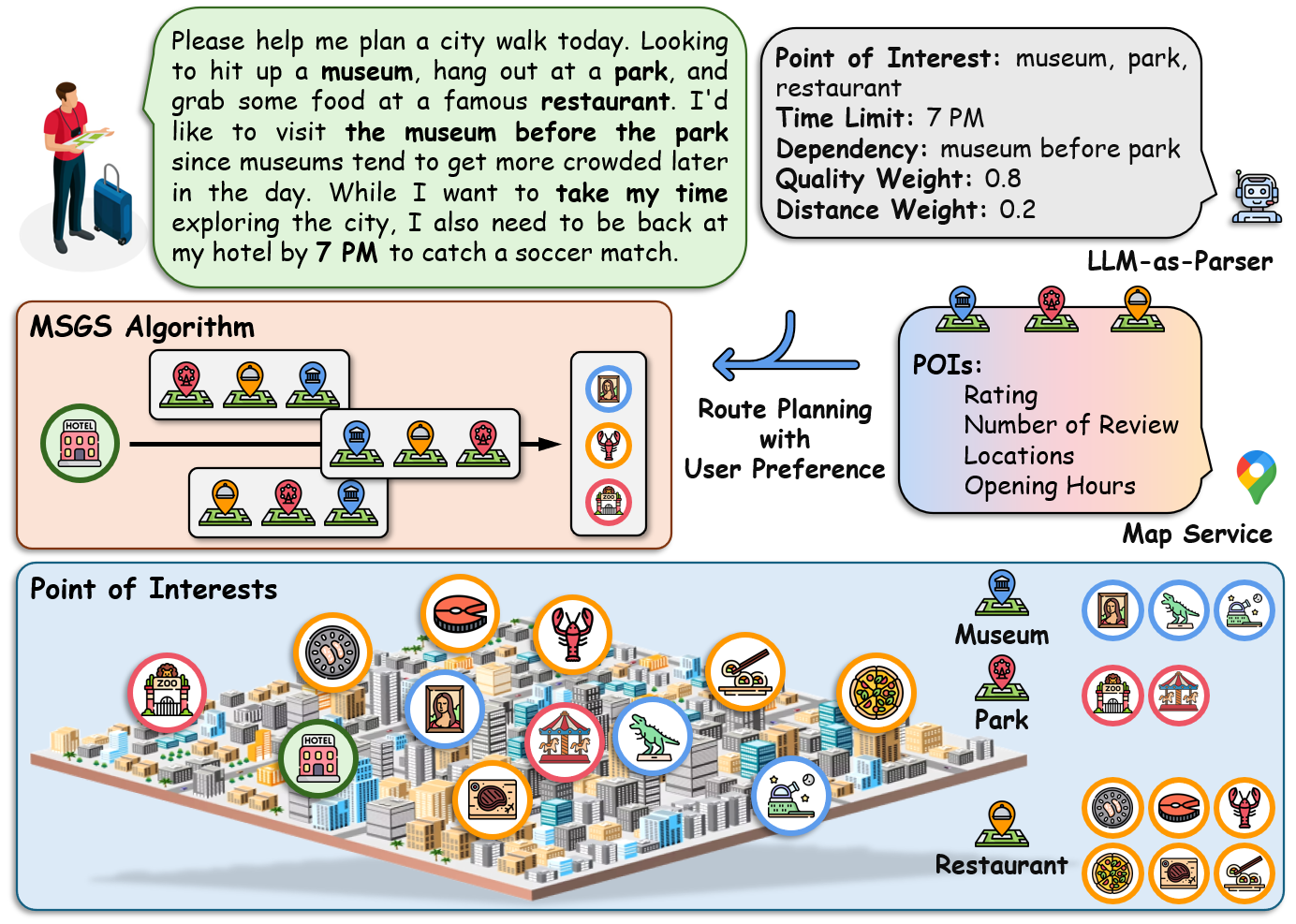}
    \caption{Overview of our \Name system. LLM-as-Parser processes user queries to extract key information and user preferences, which \AlgName then leverages to perform route planning.}
    \label{fig:Overview}
\end{figure}

While recent LLM literature has explored spatio-temporal comprehension and route optimization capabilities, most studies conduct only simplified evaluations in toy grid environments \cite{aghzal2023can, fatemi2024test}. These works typically focus on a single objective, such as minimizing distance \cite{meng2024llm, xiao2023llm}, while other approaches focus solely on constraint satisfaction without considering distance or POI quality \cite{xie2024travelplanner, hao2024large, zhang2024meta}. A typical use case illustrates the complexity in real-world scenario that such works fails to account for: LLM needs to process qualities (e.g., ratings) for multiple POIs and make weighted trade-offs between qualities and distances based on user preferences, while considering user time limits, POI opening hours, and task sequential dependencies (e.g., purchasing flowers at a florist before visiting a relative in the hospital). Furthermore, the context length grows linearly with the number of POIs, for example, 100 POIs can require up to 20,000 tokens. In practice, these challenges are amplified by the highly heterogeneous nature of user queries, which can originate from anywhere in the world at any time. Consequently, this combination of complex information, inherently multi-objective human expectations, and diverse user preferences poses a critical challenge: \textit{How can LLMs perform route planning based on user preferences, especially when faced with large amounts of heterogeneous and complex POI information?}

In this paper, we introduce \Name, a novel system that combines an \textit{LLM-as-Parser} with a novel multi-objective optimization algorithm for user preference-based route planning, as shown in Figure \ref{fig:Overview}. This architecture addresses fundamental limitations in existing approaches that rely solely on \textit{LLM-as-Agent}. \Name supports the following capabilities: (i) trade-offs between POI quality and distance according to user preferences, (ii) adaptively tunes objective weights on a per-query basis, (iii) maximization of task completion rate, and (iv) guarantee of constraints, including user time limits, POI opening hours, and task dependencies. The \Name system exhibits inherent scalability, extending beyond these specific settings to accommodate various use cases. Recall that the unique challenge in our scenario is that user locations and other information are not known a priori; each user query is highly heterogeneous and unpredictable, potentially originating from anywhere in the world at any time. To address these distinct challenges, we propose a novel solution that employs LLM-as-Parser to interpret human language, coupled with a multi-step graph construction with iterative search (\AlgName) algorithm. Our main contributions can be summarized as follows.
\begin{itemize}[leftmargin=*, itemsep=0pt]
    \item We develop the \Name system that performs route planning by interpreting human language and user preferences on a per-query basis. The system enables conversational interaction for real-time preference interpretation and error correction, while trading off multiple user objectives and adhering to various operational constraints throughout the planning process.
    \item We present the \AlgName algorithm for multi-step multi-objective optimization: first ensuring adherence to constraints including user time limits, POI opening hours, and task dependencies, then prioritizing task completion rate, followed by quality-distance trade-off optimization based on user preferences.
    \item We conduct extensive experiments on 1,000 routing prompts across 14 countries and 27 cities with heterogeneous POI distributions, evaluating 10+ LLMs with both vanilla and chain-of-thought (CoT) prompting strategies.
    \item We benchmark both our \Name system and LLM-as-Agent solutions through comprehensive experiments, demonstrating \Name's consistent and substantial advantages in handling multiple routing objectives (e.g., task completion, quality, distance) and constraints (e.g., user time limit, task dependencies, opening hours), while maintaining superior runtime efficiency.
\end{itemize}

\begin{figure*}[t]
    \centering
    \includegraphics[width=1\linewidth]{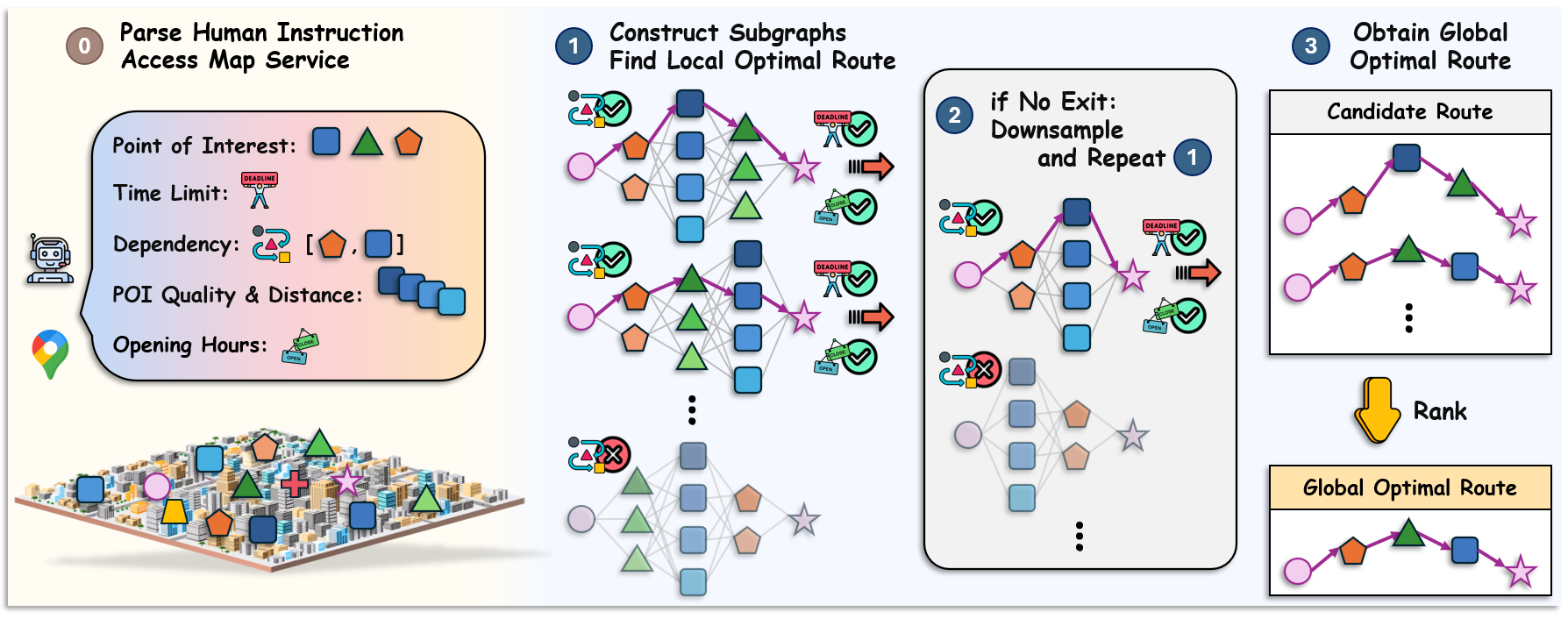}
    \caption{System architecture of \Name system and \AlgName algorithm. After the LLM-as-Parser interprets the user query and retrieves POI information from the map service, the \AlgName algorithm is used to identify the optimal route. Dependency constraints are verified before subgraph construction, while time limits and opening hours are validated after obtaining local optimal paths.}
    \label{fig:Methodology}
\end{figure*}

\section{Related Work}

\subsection{LLM for Planning}

Recent advances have explored the integration of LLMs into various planning domains \cite{dagan2023dynamic, valmeekam2023planning, wu2024can}. These works leverage LLMs' natural language understanding capabilities to interpret user requirements while addressing the spatio-temporal characteristics inherent in planning tasks. In travel planning applications, LLMs are employed to design multi-day itineraries across different cities while considering multiple constraints such as budget limitations, inter-city transportation, and attraction diversity \cite{xie2024travelplanner, tang2024synergizing, hao2024large, singh2024personal, wu2024toolplanner, ju2024globe}. For more granular planning tasks, such as path planning in sandbox environments with defined start points, endpoints, and obstacles, LLMs demonstrate capability in environment comprehension and optimal path generation \cite{xiao2023llm, meng2024llm}. In the domains of robotics and autonomous driving, planning tasks typically involve multiple obstacles in real-world environments, with visual perception playing a crucial role in the capture of environmental information \cite{shah2023lm, sinha2024real, kannan2024smart, han2024llm}.

\subsection{Route Planning Algorithm}

Route planning algorithms form the basis of autonomous navigation, with traditional approaches like Dijkstra, A*, D*, and D*-Lite known for optimality and completeness \cite{koenig2002improved, bast2016route}. These classical methods, especially Dijkstra's algorithm \cite{haeupler2024universal}, perform well on simple graphs. However, they may face computational challenges when scaling to large-scale maps. Recent research has explored graph reinforcement learning to address these limitations \cite{yu2021deep, xing2022graph}. However, existing approaches face limitations in their reliance on large-scale training data and out-of-distribution generalization, while lacking adaptive mechanisms to optimize routing on a per-query basis and balance multiple objectives. Furthermore, with human language as the desired system input, end-to-end learning and inference becomes challenging \cite{chen2024exploring}, while non-end-to-end approaches suffer from information loss and inability for joint training \cite{jin2024large, fan2024graph}. Given these challenges and the unique characteristics of our research problem, including highly heterogeneous graphs, limited node and edge counts (tens to hundreds), and diverse user preferences, we propose the combination of LLM-as-Parser and the \AlgName algorithm as a promising solution.

\section{Methodology}

\subsection{\Name System Overview}

In the \Name system, a user interacts with the LLM to derive target POI types and constraints from a language instruction. Figure \ref{fig:Overview} shows a classic scenario, which involves a user query:
\begin{Quote}
Please help me plan a city walk today. Looking to hit up a \textbf{museum}, hang out at a \textbf{park}, and grab some food at a famous \textbf{restaurant}. I'd like to visit \textbf{the museum before the park} since museums tend to get more crowded later in the day. While I want to \textbf{take my time} exploring the city, I also need to be back at my hotel by \textbf{7 PM} to catch a soccer match.
\end{Quote}
\noindent This query contains multiple pieces of information: (i) POI types: museum, park, and restaurant, (ii) user preference: prioritize quality, (iii) task dependency: museum before park, and (iv) user time limit: before 7 PM. Based on the POI types, we search existing map services, such as Google Maps, to obtain candidate POI information, including (a) ratings, (b) number of reviews, (c) geographical locations, and (d) opening hours. Subsequently, we implement the \AlgName algorithm, as shown in Figure \ref{fig:Methodology}, which utilizes a multi-step approach for graph construction, followed by route planning using search algorithms (e.g., Dijkstra) to ensure optimal solutions across multiple objectives.

\let\oldnl\nl
\newcommand{\nlnonumber}{\renewcommand{\nl}{\let\nl\oldnl}}

\begin{algorithm}[t]
\setlength{\hsize}{\linewidth}
\SetAlCapFnt{\small}
\caption{\small POI Graph Construction}
\label{alg:Graph}
\small
\nlnonumber \hspace{-10pt} \textbf{Input from LLM-as-Parser:} POI types ($\mathcal{Y}$) and user preferences on POI quality ($\mathbf{a}$) and distance ($\mathrm{b}$)

\nlnonumber \hspace{-10pt} \textbf{Input from Map Service:} POI attributes ($\mathcal{X}$)

\nlnonumber \hspace{-10pt} \textbf{Output:} POI Graph $\mathcal{G}$

Initialize empty node set $\mathcal{V} \leftarrow \emptyset$ \\
\For{each $y \in \mathcal{Y}$}{
    Interact with map services to retrieve POI data $X_y$ \\
    \For{each retrieved POI $x \in X_y$}{
        Add node $v$ with attributes $x$ to $\mathcal{V}$ \\
    }
}

Initialize empty edge set $\mathcal{E} \leftarrow \emptyset$ \\
\For{each pair of nodes $v_i, v_j \in \mathcal{V}$ where $i \neq j$}{
    Calculate edge weight {\scriptsize $\mathrm{w}_{i,j} = \mathbf{a} \cdot \mathcal{X}^+(v_j) - \mathrm{b} \cdot \delta(v_i, v_{j})$} \\
    Add edge $(v_i, v_j)$ with weight $\mathrm{w}_{i,j}$ to $\mathcal{E}$ \\
}

Return graph $\mathcal{G} = (\mathcal{V}, \mathcal{E}, X, \mathcal{Y})$
\end{algorithm}

\begin{algorithm}[t]
\setlength{\hsize}{0.95\linewidth}
\SetAlCapFnt{\small}
\caption{\small Multi-Step Graph Construction with Iterative Search (\AlgName)}
\label{alg:MSGS}
\small
\nlnonumber \hspace{-10pt} \textbf{Input from LLM-as-Parser:} POI types ($\mathcal{Y}$), user preferences on POI quality ($\mathbf{a}$) and distance ($\mathrm{b}$), user time limit ($T_{\text{user}}$), and a set of task dependency requirements ($\mathcal{D}_{\text{user}}$)

\nlnonumber \hspace{-10pt} \textbf{Input from Map Service:} POI attributes ($\mathcal{X}$)

\nlnonumber \hspace{-10pt} \textbf{Output:} Optimal route ($\xi^\ast$) 

Initialize full graph $\mathcal{G} = (\mathcal{V}, \mathcal{E}, \mathcal{X}, \mathcal{Y})$ \hfill $\rhd$ Alg. \ref{alg:Graph} \\
Initialize optimal route $\xi^\ast \leftarrow \emptyset$ \\
\For{each set $\mathbf{y} \subseteq \mathcal{Y}$}{
    \For{permutation $\vec{\mathbf{y}}$ of $\mathbf{y}$}{
        \If{$\vec{\mathbf{y}}$ satisfies $\mathcal{D}_{\text{user}}$}{
            Construct a subgraph $\mathcal{G}_{\vec{\mathbf{y}}}$ \\
            Search on $\mathcal{G}_{\vec{\mathbf{y}}}$ to find its local optimal route $\widetilde{\xi^\ast}$ \\
            \If{$\widetilde{\xi^\ast}$ satisfies $T_{\text{user}}$ and POI opening hours $T_{v_i}, \; \forall v_i \in \widetilde{\xi^\ast}$}{
                $\xi^\ast \leftarrow \arg\max_{\xi \in \{\xi^\ast, \widetilde{\xi^\ast}\}} \mathcal{O}_{\mathrm{(ii)}}(\xi)$ \\
            }
        }
    }
    \If{$\exists \; \xi^\ast$}{
        Early stop \hfill $\rhd$ Route Found
    }
}
\Else{$\xi^\ast \leftarrow [\text{start point, end point}]$ \hfill $\rhd$ No Route Found}
\end{algorithm}

\subsection{Graph Construction} 

After acquiring POI data through map service interactions, we employ Algorithm \ref{alg:Graph} to construct a POI graph $\mathcal{G} = (\mathcal{V}, \mathcal{E}, \mathcal{X}, \mathcal{Y})$, wherein $\mathcal{V}$ represents the set of nodes (i.e., discrete POIs), $\mathcal{E}$ denotes the set of edges (delineating potential routes between POIs), $\mathcal{X}$ encompasses the node attribute set (e.g., ratings, number of reviews, geographical locations, and opening hours), and $\mathcal{Y}$ represents node types (e.g., hospital, supermarket). Notably, node types (a.k.a. POI types) are defined based on tasks in the user query and can be further refined beyond the previously mentioned types (e.g., a user's task might specify Walmart and Sam's Club). Additionally, another scenario involves geographical types, such as when users want to visit the Hollywood area without being restricted to a specific POI. Recall the unique challenge in our scenario of constructing graphs from natural language instructions, each query results in a highly heterogeneous POI graph $\mathcal{G}$, varying not only in attributes $\mathcal{X}$ and types $\mathcal{Y}$ but also in the number of nodes $|\mathcal{V}|$ and the number of edges $|\mathcal{E}|$.

\subsection{Multi-Objective and User Preferences}

Users inherently have multiple objectives, and their preference priorities vary across different queries. For example, during travel, users emphasize ratings of attractions and renowned restaurants, while during busy times, they prioritize shorter routes. Let $\xi = (v_0, v_1, ..., v_N)$ denote a route in graph $\mathcal{G}$, where $v_i \in \mathcal{V}$ and $(v_i, v_{i+1}) \in \mathcal{E}$. Mathematically, we define the multi-step multi-objective optimization problem as follows:
\begin{equation}
\resizebox{\linewidth}{!}{$\displaystyle
\begin{aligned}
\mathcal{O}_{\mathrm{(i)}} := & \max_{\xi \in \Omega} \left(\frac{|\left\{\mathcal{Y}(v_i) | v_i \in \xi\right\}|}{|\mathcal{Y}|}\right), \\
\mathcal{O}_{\mathrm{(ii)}} := & \max_{\xi \in \Omega} \left(\sum_{i=0}^{N} \mathbf{a} \cdot \mathcal{X}^+(v_{i+1}) - \mathrm{b} \sum_{i=0}^{N-1} \delta(v_i, v_{i+1})\right), \\
\mathrm{s.t.} \;
& \sum_{i=0}^{N} \mathcal{X}^T(v_i) + \sum_{i=0}^{N-1} \delta(v_i, v_{i+1}) \leq T_{\text{user}}, \\
& \sum_{\tau=0}^{i} \mathcal{X}^T(v_\tau) + \sum_{\tau=0}^{i-1} \delta(v_\tau, v_{\tau+1}) \leq T_{v_i}, \; \forall v_i \in \xi, \\
& v_i \prec v_j, \quad \forall (v_i, v_j) \in \mathcal{D}_{\text{user}},
\end{aligned}
$}
\label{eq:optimization}
\end{equation}
where $\Omega$ represents all possible routes in graph $\mathcal{G}$, $\delta(v_i, v_{i+1})$ denotes the travel time between consecutive POIs, and $\mathcal{X}^T(v_i)$ represents the duration spent at each POI. This multi-step multi-objective optimization: $\mathcal{O}_{\mathrm{(i)}}$ maximizes task completion rate by optimizing the coverage of desired POI types, and $\mathcal{O}_{\mathrm{(ii)}}$ optimizes the trade-off between POI quality and route distance. We cumulatively optimize positive attributes $\mathcal{X}^+$ (such as ratings and number of reviews) of each POI visited along the route, while minimizing the total distance traveled. The weight vector $\mathbf{a}$ and the scalar weight $\mathrm{b}$ represent the positive importance of POI attributes and the route distance, respectively, which are dynamically adjusted based on user preferences expressed in natural language. The optimization is constrained by user time limit $T_{\text{user}}$, opening hours for each POI $T_{v_i}$, and task sequential dependencies $\mathcal{D}_{\text{user}}$.

\subsection{\AlgName Algorithm}

Next, we present our proposed \AlgName algorithm, which focuses on reconstructing $\mathcal{G}$ to obtain the optimal route $\xi^\ast$. In real-world scenarios, $\mathcal{G}$ typically exhibits a fully connected topology, since humans can inherently travel between two locations. However, this would lead to search failure, as standard search algorithms (e.g., Dijkstra) would likely proceed directly to the destination, failing to maintain task completion rates. Therefore, we propose a novel \AlgName algorithm to perform multi-step multi-objective optimization. To satisfy $\mathcal{O}_{\mathrm{(i)}}$ task completion rate, we reorganize and reconstruct the original graph $\mathcal{G}$ into multiple directed subgraphs, then employ a search strategy (in our paper, the Dijkstra algorithm) to obtain the optimal path. Notably, due to the joint effect of multiple constraints, we traverse all combinations of $\mathcal{Y}$ and its subsets to ensure constraint satisfaction. To address $\mathcal{O}_{\mathrm{(ii)}}$ trade-off between POI quality and route distance based on user preferences, we use LLM-as-Parser to interpret the user language and derive the weight vector $\mathbf{a}$ and the scalar weight $\mathrm{b}$. The detailed implementation of our \Name system is outlined in Algorithm \ref{alg:MSGS}.

\section{Experiments}

We aim to address two key research questions in our experiments: (RQ1) \textit{Does our \Name system outperform pure LLM solutions (i.e., LLM-as-Agent) in route planning tasks?} (RQ2) \textit{How effectively can LLM-as-Parser extract accurate information from human instructions?}

\subsection{Dataset Generation}

\textbf{Human Instruction.} To evaluate the performance of LLM-as-Agent and LLM-as-Parser, we construct a synthetic dataset called Human Instructions with Preferences for route Planning (HIPP). The dataset comprises 1,000 evaluation samples, each with heterogeneous settings. An example from the HIPP dataset are provided in Appendix \ref{appendix:HIPP}. The HIPP dataset generation process consists of three main steps:
\begin{enumerate}[leftmargin=*]
    \item \textbf{Synthetic Label:} We randomly sample one to five POI types from a predefined taxonomy comprising shopping malls, supermarkets, pharmacies, banks, and libraries, utilizing a discrete uniform distribution. Temporal constraints are incorporated with probability 0.3 by generating stochastic time limits between 17:00 and 24:00, sampled from a discrete uniform distribution. Furthermore, task dependencies are established with probability 0.3 between POI categories via independent Bernoulli trials conducted for each consecutive POI pair, thereby modeling sequential task relationships. The user preference weights for the quality and distance of POI are generated using a discrete uniform distribution over the quantized interval \{0, 0.1, 0.2, ..., 1.0\}, with the constraint that their sum equals unity to ensure proper normalization. POI operational hours are derived from authentic real-world data obtained through the Google Maps API.
    \item \textbf{Human Instruction (Synthetic Data):} We leverage GPT-4o \cite{GPT4o} to generate natural instructions based on these synthetic labels through CoT prompting \cite{wei2022chain, kojima2022large}. These instructions are designed to be natural and implicit, deliberately excluding explicit numerical preference weights.
    \item[3a.] \textbf{LLM-as-Agent:} We employ LLMs to comprehend human instructions and process map data, including POI geographical locations, ratings, review counts, and other relevant attributes. To address the token length limitations due to numerous POIs, we sample 10 specific POIs for each POI type as input to LLMs.
    \item[3b.] \textbf{LLM-as-Parser with \AlgName:} We implement LLMs to interpret, identify, and estimate parameters from these human instructions, generating estimations in the same JSON format as the synthetic label for evaluation purposes. 
\end{enumerate}

\noindent\textbf{Use Case Scenario.}
We utilize the Google Places API \footnote{\url{https://developers.google.com/maps/documentation/places/web-service/overview}} to retrieve detailed POI information, including geographical locations, ratings, number of reviews, opening hours, etc. Our evaluation scenario simulates a common use case: a student traveling from an airport to a university campus while completing intermediate tasks, such as shopping at a supermarket. To ensure generalizability, we conduct experiments across 27 major cities in 14 countries. Detailed information on the selected cities and their information can be found in Appendix \ref{appendix:map service}.

\newcommand{\CoT}[1]{\scriptsize (CoT)}
\newcommand{\com}[1]{\tiny$\pm$#1}

\begin{table*}[t]
\centering 
\resizebox{\linewidth}{!}{
\begin{tabular}{ll|cccc|ccc}
\toprule
\multirow{2.1}{*}{\textbf{Method}} & \multirow{2.1}{*}{\textbf{Model}} & \textbf{Rating} & \textbf{Number of} & \textbf{Length} & \textbf{Task Completion} & \multicolumn{3}{c}{\textbf{Constraint Violation} ($\downarrow$)} \\
& & ($\uparrow$) & \textbf{Review} ($\uparrow$) & (km) ($\downarrow$) & \textbf{Rate} (\%) ($\uparrow$) & \textbf{Time Limit} (hrs) & \textbf{Dependency} (\%) &  \textbf{Opening Hours} (\%) \\
\midrule
\multicolumn{2}{l|}{\Name w/o LLM-as-Parser} & \cellcolor{Yes} 4.30 & \cellcolor{Yes} 5936 & \cellcolor{Yes} 29.78 & \cellcolor{Yes} 96.69 & \cellcolor{Yes} 0.00 & \cellcolor{Yes} 0.00 & \cellcolor{Yes} 0.00 \\
\multicolumn{2}{l|}{\Name w/o \AlgName} & \cellcolor{Nooo} 1.66 & \cellcolor{Yes} 5674 & \cellcolor{No} 42.45 & \cellcolor{Nooo} 17.29 & \cellcolor{Yes} 0.00 & \cellcolor{No} 41.23 & \cellcolor{No} 1.50 \\

\multicolumn{2}{l|}{SMT Solver \cite{hao2024large}} & \cellcolor{Nooo} 1.22 & \cellcolor{No} 158 & \cellcolor{No} 34.36 & \cellcolor{Nooo} 13.90 & \cellcolor{No} 0.28 & \cellcolor{No} 2.51 & \cellcolor{No} 11.10 \\
\multicolumn{2}{l|}{SMT Solver v2} & \cellcolor{Nooo} 1.98 & \cellcolor{No} 491 & \cellcolor{No} 36.71 & \cellcolor{Yes} 100.00 & \cellcolor{Yes} 0.00 & \cellcolor{No} 60.36 & \cellcolor{No} 22.40 \\

\midrule
\Name & Phi-3-mini & \cellcolor{Nooo} 2.19 & \cellcolor{Yes} 3459 & \cellcolor{Yes} 29.58 & \cellcolor{Nooo} 48.44 & \cellcolor{Yes} 0.00 & \cellcolor{Yes} 0.00 & \cellcolor{Yes} 0.00 \\
\cellcolor{Yes}\Name & \cellcolor{Yes}Phi-3-mini \CoT & \cellcolor{Yes} 4.40 & \cellcolor{Yes} 6910 & \cellcolor{Yes} 29.90 & \cellcolor{Yes} 96.65 & \cellcolor{Yes} 0.00 & \cellcolor{Yes} 0.00 & \cellcolor{Yes} 0.00 \\

\midrule

LLM-as-Agent & Phi-3.5-mini & \cellcolor{No} 3.85 & \cellcolor{Yes} 1514 & \cellcolor{No} 249.50 & \cellcolor{Yes} 92.72 & \cellcolor{Nooo} 12.14 & \cellcolor{No} 56.04 & \cellcolor{No} 96.30 \\
LLM-as-Agent & Phi-3.5-mini \CoT & \cellcolor{No} 3.85 & \cellcolor{Yes} 1510 & \cellcolor{No} 247.33 & \cellcolor{Yes} 92.72 & \cellcolor{Nooo} 11.88 & \cellcolor{No} 56.04 & \cellcolor{No} 96.50 \\
\Name & Phi-3.5-mini & \cellcolor{No} 3.69 & \cellcolor{Yes} 5471 & \cellcolor{Yes} 29.73 & \cellcolor{No} 81.60 & \cellcolor{Yes} 0.00 & \cellcolor{Yes} 0.00 & \cellcolor{Yes} 0.00 \\
\cellcolor{Yes}\Name & \cellcolor{Yes}Phi-3.5-mini \CoT & \cellcolor{Yes} 4.38 & \cellcolor{Yes} 6616 & \cellcolor{Yes} 29.59 & \cellcolor{Yes} 96.47 & \cellcolor{Yes} 0.00 & \cellcolor{Yes} 0.00 & \cellcolor{Yes} 0.00 \\

\midrule

LLM-as-Agent & LLaMA-3.2-3B & \cellcolor{Yes} 4.01 & \cellcolor{Yes} 2222 & \cellcolor{No} 59.29 & \cellcolor{Nooo} 50.10 & \cellcolor{No} 0.13 & \cellcolor{No} 9.57 & \cellcolor{No} 34.40 \\
LLM-as-Agent & LLaMA-3.2-3B \CoT & \cellcolor{Yes} 4.05 & \cellcolor{Yes} 2409 & \cellcolor{No} 56.50 & \cellcolor{Nooo} 49.78 & \cellcolor{No} 0.05 & \cellcolor{No} 9.11 & \cellcolor{No} 27.00 \\
\Name & LLaMA-3.2-3B & \cellcolor{Yes} 4.00 & \cellcolor{Yes} 5860 & \cellcolor{No} 30.21 & \cellcolor{Yes} 91.53 & \cellcolor{No} 0.09 & \cellcolor{Yes} 0.00 & \cellcolor{Yes} 0.00 \\
\Name & LLaMA-3.2-3B \CoT & \cellcolor{Yes} 4.06 & \cellcolor{Yes} 6788 & \cellcolor{No} 30.57 & \cellcolor{No} 88.48 & \cellcolor{No} 0.14 & \cellcolor{Yes} 0.00 & \cellcolor{Yes} 0.00 \\

\midrule

LLM-as-Agent & LLaMA-3.1-8B & \cellcolor{No} 3.98 & \cellcolor{Yes} 2988 & \cellcolor{No} 72.98 & \cellcolor{Nooo} 54.92 & \cellcolor{Noo} 0.53 & \cellcolor{No} 7.29 & \cellcolor{No} 58.90 \\
LLM-as-Agent & LLaMA-3.1-8B \CoT & \cellcolor{No} 3.89 & \cellcolor{Yes} 2878 & \cellcolor{No} 66.85 & \cellcolor{Nooo} 51.16 & \cellcolor{Noo} 0.40 & \cellcolor{No} 3.19 & \cellcolor{No} 46.00 \\
\Name & LLaMA-3.1-8B & \cellcolor{No} 3.87 & \cellcolor{Yes} 5592 & \cellcolor{No} 30.32 & \cellcolor{No} 90.00 & \cellcolor{No} 0.12 & \cellcolor{Yes} 0.00 & \cellcolor{Yes} 0.00 \\
\Name & LLaMA-3.1-8B \CoT & \cellcolor{Yes} 4.07 & \cellcolor{Yes} 6449 & \cellcolor{No} 30.46 & \cellcolor{Yes} 90.24 & \cellcolor{No} 0.26 & \cellcolor{Yes} 0.00 & \cellcolor{Yes} 0.00 \\

\midrule

LLM-as-Agent & Mistral-7B & \cellcolor{Yes} 4.09 & \cellcolor{Yes} 3313 & \cellcolor{No} 53.67 & \cellcolor{Nooo} 46.46 & \cellcolor{No} 0.01 & \cellcolor{No} 3.42 & \cellcolor{No} 19.90 \\
LLM-as-Agent & Mistral-7B \CoT & \cellcolor{Yes} 4.12 & \cellcolor{Yes} 3623 & \cellcolor{No} 53.60 & \cellcolor{Nooo} 46.78 & \cellcolor{Yes} 0.00 & \cellcolor{No} 3.64 & \cellcolor{No} 21.00 \\
\Name & Mistral-7B & \cellcolor{Yes} 4.16 & \cellcolor{Yes} 5819 & \cellcolor{Yes} 29.85 & \cellcolor{Yes} 94.58 & \cellcolor{No} 0.14 & \cellcolor{Yes} 0.00 & \cellcolor{Yes} 0.00 \\
\cellcolor{Yes}\Name & \cellcolor{Yes}Mistral-7B \CoT & \cellcolor{Yes} 4.30 & \cellcolor{Yes} 6071 & \cellcolor{Yes} 29.56 & \cellcolor{Yes} 96.44 & \cellcolor{Yes} 0.00 & \cellcolor{Yes} 0.00 & \cellcolor{Yes} 0.00 \\

\midrule

LLM-as-Agent & Gemma-2-2B & \cellcolor{No} 3.78 & \cellcolor{Yes} 1529 & \cellcolor{No} 59.45 & \cellcolor{Nooo} 34.62 & \cellcolor{No} 0.05 & \cellcolor{No} 6.83 & \cellcolor{No} 27.80 \\
LLM-as-Agent & Gemma-2-2B \CoT & \cellcolor{No} 3.77 & \cellcolor{Yes} 1543 & \cellcolor{No} 57.54 & \cellcolor{Nooo} 33.62 & \cellcolor{No} 0.04 & \cellcolor{No} 5.69 & \cellcolor{No} 23.70 \\
\Name & Gemma-2-2B & \cellcolor{Yes} 4.40 & \cellcolor{Yes} 6909 & \cellcolor{No} 30.34 & \cellcolor{Yes} 96.03 & \cellcolor{Yes} 0.00 & \cellcolor{Yes} 0.00 & \cellcolor{Yes} 0.00 \\
\cellcolor{Yes}\Name & \cellcolor{Yes}Gemma-2-2B \CoT & \cellcolor{Yes} 4.36 & \cellcolor{Yes} 6552 & \cellcolor{Yes} 29.68 & \cellcolor{Yes} 95.46 & \cellcolor{Yes} 0.00 & \cellcolor{Yes} 0.00 & \cellcolor{Yes} 0.00 \\

\midrule

LLM-as-Agent & Gemma-2-9B & \cellcolor{Yes} 4.09 & \cellcolor{Yes} 3585 & \cellcolor{No} 53.89 & \cellcolor{Nooo} 53.34 & \cellcolor{Yes} 0.00 & \cellcolor{No} 2.05 & \cellcolor{No} 21.80 \\
LLM-as-Agent & Gemma-2-9B \CoT & \cellcolor{Yes} 4.10 & \cellcolor{Yes} 3506 & \cellcolor{No} 54.15 & \cellcolor{Nooo} 53.60 & \cellcolor{Yes} 0.00 & \cellcolor{No} 1.59 & \cellcolor{No} 20.80 \\
\Name & Gemma-2-9B & \cellcolor{Yes} 4.05 & \cellcolor{Yes} 5077 & \cellcolor{Yes} 29.53 & \cellcolor{Yes} 93.76 & \cellcolor{No} 0.25 & \cellcolor{Yes} 0.00 & \cellcolor{Yes} 0.00 \\
\Name & Gemma-2-9B \CoT & \cellcolor{No} 3.95 & \cellcolor{Yes} 4823 & \cellcolor{Yes} 29.52 & \cellcolor{Yes} 93.91 & \cellcolor{No} 0.33 & \cellcolor{Yes} 0.00 & \cellcolor{Yes} 0.00 \\

\midrule

\Name & GPT-3.5 & \cellcolor{Yes} 4.23 & \cellcolor{Yes} 6957 & \cellcolor{No} 30.20 & \cellcolor{Yes} 92.48 & \cellcolor{No} 0.30 & \cellcolor{Yes} 0.00 & \cellcolor{Yes} 0.00 \\
\Name & GPT-3.5 \CoT & \cellcolor{Yes} 4.20 & \cellcolor{Yes} 6660 & \cellcolor{No} 30.16 & \cellcolor{Yes} 91.81 & \cellcolor{Noo} 0.34 & \cellcolor{Yes} 0.00 & \cellcolor{Yes} 0.00 \\

\midrule

\Name & GPT-4o-mini & \cellcolor{Yes} 4.24 & \cellcolor{Yes} 6249 & \cellcolor{Yes} 29.68 & \cellcolor{Yes} 94.03 & \cellcolor{No} 0.25 & \cellcolor{Yes} 0.00 & \cellcolor{Yes} 0.00 \\
\Name & GPT-4o-mini \CoT & \cellcolor{Yes} 4.24 & \cellcolor{Yes} 6172 & \cellcolor{Yes} 29.68 & \cellcolor{Yes} 94.03 & \cellcolor{No} 0.30 & \cellcolor{Yes} 0.00 & \cellcolor{Yes} 0.00 \\
\cellcolor{Yes}\Name & \cellcolor{Yes}GPT-4o & \cellcolor{Yes} 4.26 & \cellcolor{Yes} 5720 & \cellcolor{Yes} 29.70 & \cellcolor{Yes} 96.02 & \cellcolor{Yes} 0.00 & \cellcolor{Yes} 0.00 & \cellcolor{Yes} 0.00 \\
\cellcolor{Yes}\Name & \cellcolor{Yes}GPT-4o \CoT & \cellcolor{Yes} 4.28 & \cellcolor{Yes} 5530 & \cellcolor{Yes} 29.73 & \cellcolor{Yes} 96.44 & \cellcolor{Yes} 0.00 & \cellcolor{Yes} 0.00 & \cellcolor{Yes} 0.00 \\

\midrule

\Name & OpenAI o1-mini & \cellcolor{Yes} 4.07 & \cellcolor{Yes} 5372 & \cellcolor{Yes} 29.69 & \cellcolor{Yes} 93.78 & \cellcolor{No} 0.31 & \cellcolor{Yes} 0.00 & \cellcolor{Yes} 0.00 \\
\Name & OpenAI o1-mini \CoT & \cellcolor{Yes} 4.09 & \cellcolor{Yes} 5348 & \cellcolor{Yes} 29.60 & \cellcolor{Yes} 93.69 & \cellcolor{No} 0.32 & \cellcolor{Yes} 0.00 & \cellcolor{Yes} 0.00 \\
\Name & OpenAI o1 & \cellcolor{Yes} 4.12 & \cellcolor{Yes} 5193 & \cellcolor{Yes} 29.54 & \cellcolor{Yes} 93.53 & \cellcolor{Noo} 0.34 & \cellcolor{Yes} 0.00 & \cellcolor{Yes} 0.00 \\
\Name & OpenAI o1 \CoT & \cellcolor{Yes} 4.11 & \cellcolor{Yes} 5161 & \cellcolor{Yes} 29.48 & \cellcolor{Yes} 93.49 & \cellcolor{Noo} 0.34 & \cellcolor{Yes} 0.00 & \cellcolor{Yes} 0.00 \\
\bottomrule 
\end{tabular}
}
\caption{Main results for route planning evaluation across different methods. Results use a color scheme where blue indicates superior and red indicates inferior performance, with detailed color threshold specifications in Appendix \ref{appendix:Evaluation Metrics}. We observe that the proposed \Name approach offers a significant compared to the LLM-as-Agent baseline across various metrics under different model settings.}
\label{Table Main Results}
\end{table*}

\subsection{Experimental Setup}

\textbf{Route Evaluation.}
For route planning, we consider the seven evaluation metrics divided into two categories: soft metrics and hard constraints. Soft metrics include rating, number of reviews, path length, and task completion rate. Hard constraints consist of user time limit, task dependency requirements, and opening hours compliance. The hard constraints represent mandatory requirements that all valid routes must satisfy, while soft metrics reflect varying degrees of route quality. 

\noindent\textbf{LLM-as-Parser Evaluation.}
We evaluate the performance of different LLMs and CoT prompting using four metrics. For POI type identification and ask dependency detection, we employ the F1 score, as they effectively distinguish between missing and superfluous elements. For user time limit extraction we use accuracy, while for user preference estimation we use similarity. Details are provided in Appendix \ref{appendix:Evaluation Metrics}. All four metrics follow a higher-is-better principle, enabling us to compute average scores across metrics.

\noindent\textbf{Implementation.}
In our experiments, we evaluate various LLMs with their corresponding CoT prompting approaches, including Phi-3-mini \cite{abdin2024phi}, Phi-3.5-mini \cite{abdin2024phi}, LLaMA-3.2-3B \cite{llama3}, LLaMA-3.1-8B \cite{llama3}, Mistral-7B-v0.3 \cite{jiang2023mistral}, Gemma-2-2b \cite{team2024gemma2}, Gemma-2-9b \cite{team2024gemma2}, GPT-3.5-turbo \cite{openai2023chatgpt}, GPT-4o-mini \cite{GPT4omini}, GPT-4o \cite{GPT4o}, OpenAI o1-mini \cite{o1}, and OpenAI o1 \cite{o1}. All experiments are conducted on an NVIDIA A100 GPU with 40 GB of memory.

\noindent\textbf{SOTA Baselines.} While numerous prompting techniques have been proposed to enhance LLMs' reasoning capabilities, such as CoT \cite{kojima2022large}, ToT \cite{yao2023tree}, GoT \cite{besta2024graph}, ReAct \cite{yao2022react}, and Reflexion \cite{shinn2024reflexion}, we primarily focus on comparing CoT with our base implementation. This choice is made as our paper focuses on comparing two fundamental paradigms (LLM-as-Agent vs. LLM-as-Parser) rather than evaluating various prompting techniques. Additionally, we observe that the limitations of LLM-as-Agent approach are primarily due to the practical constraints of handling extensive POI information, resulting in increased memory usage and inference time, rather than solely from insufficient reasoning capabilities. Most existing approaches \cite{xie2024travelplanner, singh2024personal} adopt the LLM-as-Agent paradigm, which directly employs LLMs for route planning, hence we do not include specific annotations. We also consider an SMT solver-based method for evaluation \cite{hao2024large}. For simplicity, we utilize synthetic labels and implement the SMT solver to handle three constraints: time limit, dependency, and opening hours. Given that SMT solvers are not designed for multi-step multi-objective optimization, we extend the approach to SMT solver v2, which incorporates task completion rate as an additional constraint to evaluate the solver's performance in our research scenario.

\subsection{Main Results}

\textbf{Comparison with LLM-as-Agent.} Table \ref{Table Main Results} presents a comparative analysis between our \Name system and LLM-as-Agent approach, demonstrating that \Name (LLM-as-Parser + \AlgName) consistently outperforms LLM-as-Agent across all metrics. We observe two distinct types of errors in LLM-as-Agent implementations. In the first case, certain LLMs (e.g., Phi-3.5-mini, Phi-3.5-mini (CoT)) fail to perform reasonable route planning, indiscriminately including POIs regardless of their feasibility. While this case appears to achieve high task completion rates, it significantly exceeds time limits and disregards dependency constraints. In the second case, while other LLMs demonstrate the ability to selectively incorporate POIs into routes, they struggle to effectively maximize task completion rates while satisfying multiple constraints. Although these LLMs show potential to maximize ratings and the number of reviews, they exhibit limitations in geographical distance reasoning. \Name prioritizes task completion rate and constraint satisfaction, achieving superior performance in these objectives.

\begin{table}[t]
\centering 
\resizebox{\linewidth}{!}{
\begin{tabular}{l|cccc}
\toprule
\multirow{2.1}{*}{\textbf{LLM}} & \textbf{POI} & \textbf{Time Lt.} & \textbf{Dep.} & \textbf{Preference} \\
& \textbf{F1} (\%) & \textbf{Acc.} (\%) & \textbf{F1} (\%) & \textbf{Sim.} (\%) \\
\midrule
Phi-3-mini & 51.20 & 37.42 & 26.80 & 78.58 \\
Phi-3-mini \CoT & 100.00 & 90.32 & 99.36 & 81.53 \\
Phi-3.5-mini & 85.70 & 73.87 & 74.91 & 77.81 \\
Phi-3.5-mini \CoT & 100.00 & 97.74 & 99.74 & 76.82 \\
LLaMA-3.2-3B & 98.27 & 79.35 & 95.29 & 82.08 \\
LLaMA-3.2-3B \CoT & 96.49 & 84.52 & 87.17 & 80.19 \\
LLaMA-3.1-8B & 94.50 & 86.13 & 91.48 & 78.58 \\
LLaMA-3.1-8B \CoT & 96.60 & 88.71 & 93.89 & 82.68 \\
Mistral-7B & 99.28 & 95.48 & 97.58 & 82.81 \\
Mistral-7B \CoT & 99.98 & 99.68 & 99.22 & 80.89 \\
Gemma-2-2B & 99.90 & 89.03 & 98.16 & 81.54 \\
Gemma-2-2B \CoT & 99.47 & 93.23 & 97.76 & 81.69 \\
Gemma-2-9B & 99.92 & 92.90 & 99.60 & 87.91 \\
Gemma-2-9B \CoT & 99.96 & 90.65 & 99.48 & 84.93 \\
GPT-3.5 & 100.00 & 86.13 & 99.46 & 77.99 \\
GPT-3.5 \CoT & 100.00 & 85.81 & 99.53 & 77.57 \\
GPT-4o-mini & 99.80 & 90.65 & 99.74 & 86.14 \\
GPT-4o-mini \CoT & 100.00 & 90.65 & 99.74 & 86.52 \\
GPT-4o & 99.90 & 98.06 & 99.78 & 89.35 \\
GPT-4o \CoT & 100.00 & 99.03 & 99.74 & 89.74 \\
OpenAI o1-mini & 99.97 & 91.29 & 99.66 & 88.39 \\
OpenAI o1-mini \CoT & 99.87 & 90.97 & 99.70 & 88.16 \\
OpenAI o1 & 99.90 & 90.32 & 99.74 & 90.33 \\
OpenAI o1 \CoT & 100.00 & 90.32 & 99.81 & 90.76 \\
\bottomrule 
\end{tabular}
}
\caption{Evaluation of various LLM-as-Parser models, with blue highlighting the highest average score.}
\label{Table LLM-as-Parser}
\end{table}

\noindent\textbf{Comparison with SOTA Baselines.} In our scenario, both the LLM-as-Agent solution and the SMT solver approach \cite{hao2024large} demonstrate significant limitations in generating reasonable routes. The SMT solver's limitations stem not only from its focus on constraints while neglecting the trade-offs between human objectives but also from the substantial complexity and volume of these constraints. The SMT solver tends to generate direct routes from start to end point while bypassing POIs, effectively avoiding constraint violations. However, such solutions clearly fail to meet user expectations. To address this limitation, SMT solver v2 incorporates task completion rate as an additional constraint. However, due to conflicting constraints, while SMT solver v2 satisfies task completion rate and time limit requirements, it inevitably violates dependency constraints to accommodate the former two.

\begin{figure*}[t]
    \centering
    \includegraphics[width=0.99\linewidth]{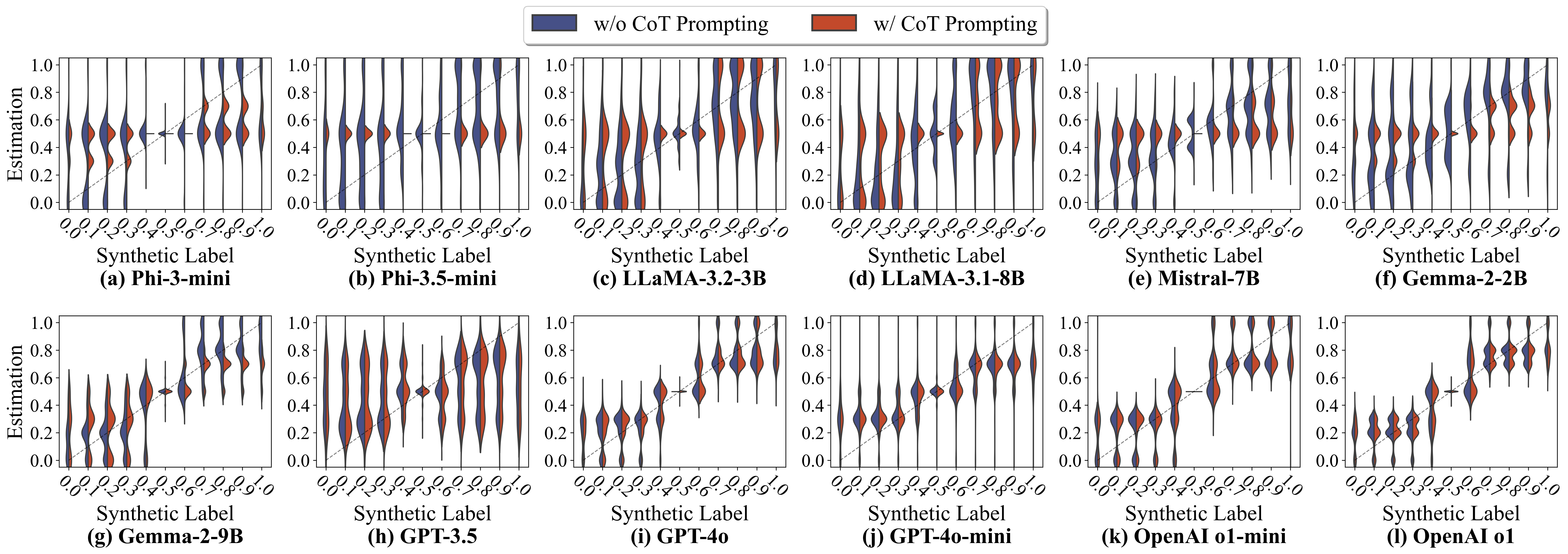}
    \caption{Impact of LLM-as-Parser on user preference weight estimation. The gray dashed line represents the ideal scenario where estimations perfectly match the synthetic labels.}
    \label{fig:Impact_Weight}
\end{figure*}

\noindent\textbf{Impact of CoT Prompting.} We observe that CoT prompting provides limited benefits for the LLM-as-Agent approach while substantially enhancing LLM-as-Parser performance. This disparity arises because CoT fundamentally aims to help LLMs emulate human step-by-step reasoning processes. However, when applied to LLM-as-Agent scenarios, neither CoT-enhanced LLMs nor humans can effectively process hundreds of POIs (in our experiments, limited to 10 POIs per type to reduce input length) along with their associated information and required computations, such as distance calculations and multi-objective trade-offs. In contrast, for \Name, CoT prompting effectively facilitates LLMs in emulating human reasoning patterns, leading to a more accurate interpretation of user instructions, particularly for Phi-3-mini and Phi-3.5-mini models. We provide a detailed analysis of CoT performance improvements for these models in Section \ref{sec:LLM-as-Parser}.

\subsection{LLM-as-Parser Evaluation}
\label{sec:LLM-as-Parser}

\noindent\textbf{Impact of Estimation Accuracy.} Table \ref{Table LLM-as-Parser} further demonstrates the reliability of LLM‐as‐Parser. We evaluate the F1 score and accuracy using synthetic samples against LLM estimations to assess their capability in extracting POIs, routes, and preference information from natural language instructions. Our results show that LLMs excel at identifying POIs, time limits, and dependencies, as these elements primarily rely on lexical cues and logical structures that can be directly parsed from the instructions. However, the accuracy in estimating preference weights is comparatively lower, as LLMs must infer numerical values from textual descriptions. Natural language rarely conveys precise numerical information, for example, when a user states ``I am in a hurry,'' the inferred distance weight could range from 0.7 to 0.9.

\begin{figure*}[t]
\centering
\subfigure[Typographical Error]{\includegraphics[width=0.32\linewidth]{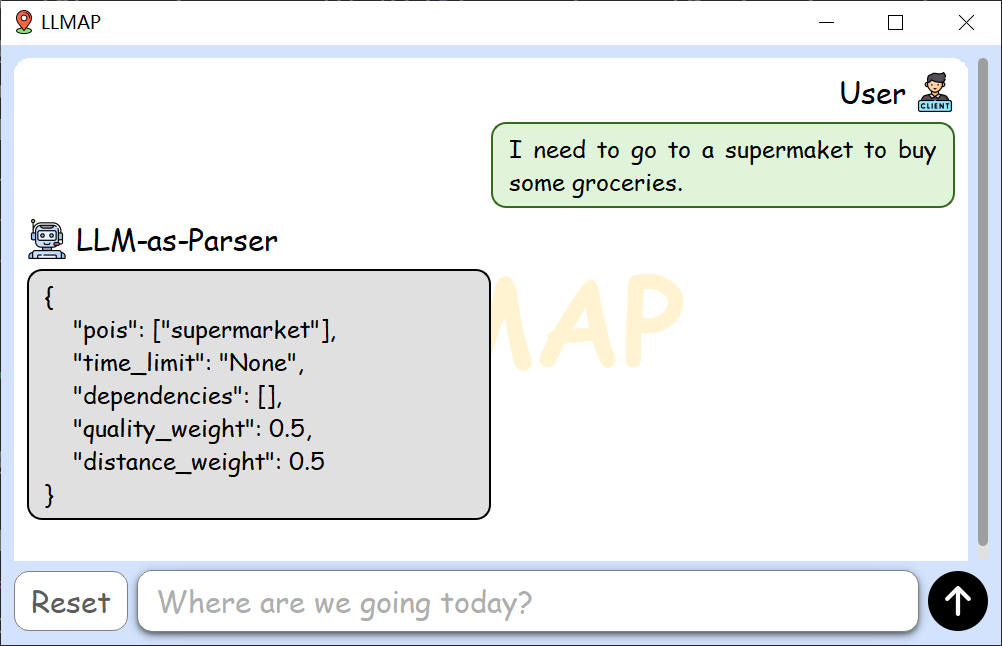}\label{fig:Error 1}}
\hfill
\subfigure[Grammar Error]{\includegraphics[width=0.32\linewidth]{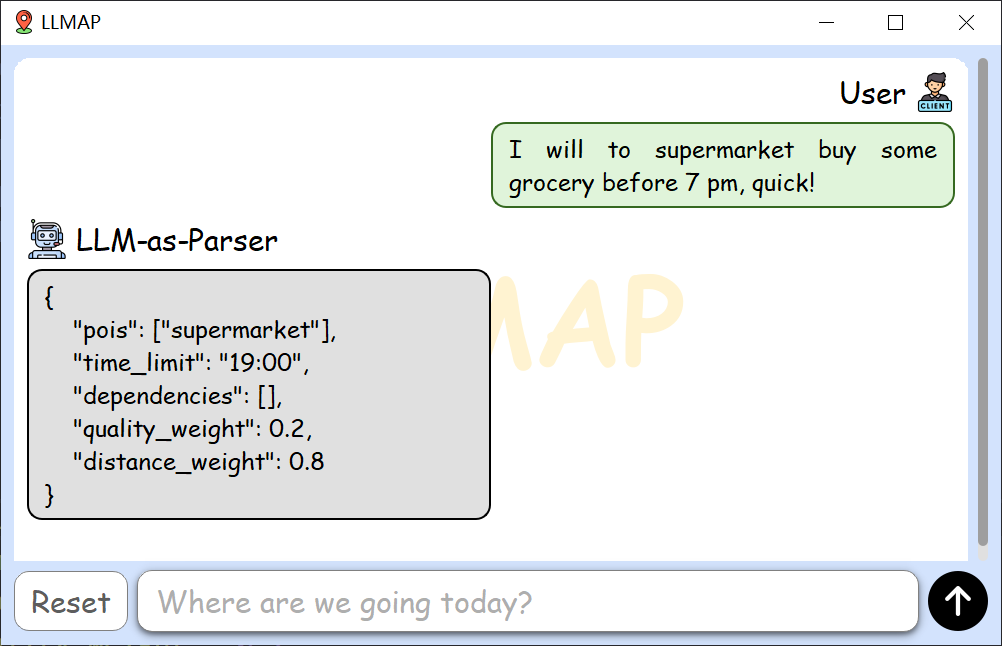}\label{fig:Error 2}}
\hfill
\subfigure[Non-Existent POI]{\includegraphics[width=0.32\linewidth]{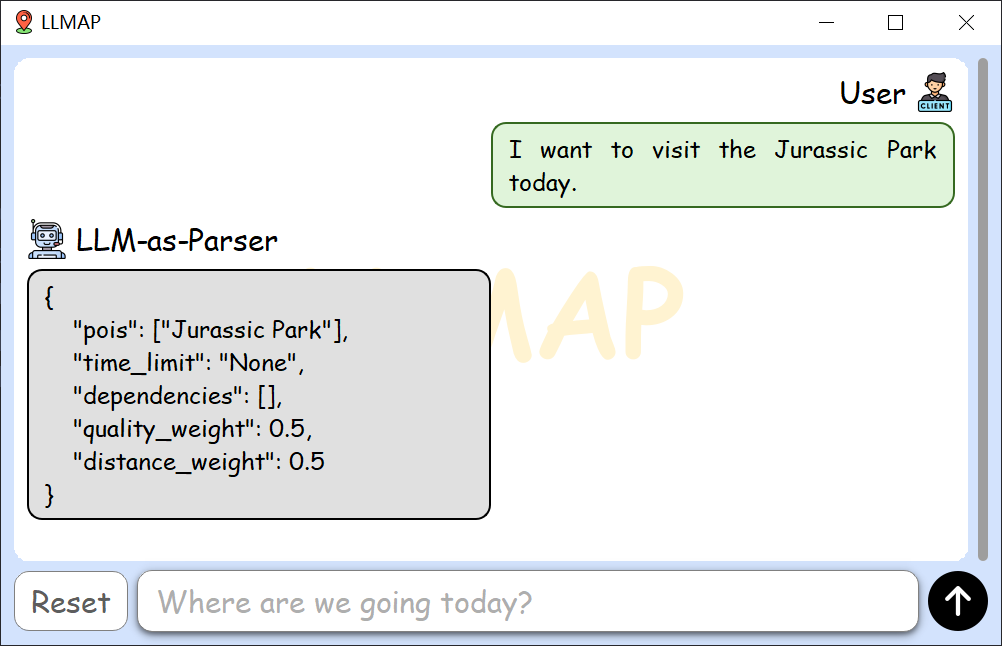}\label{fig:Error 3}}
\\
\subfigure[Vague Instruction]{\includegraphics[width=0.32\linewidth]{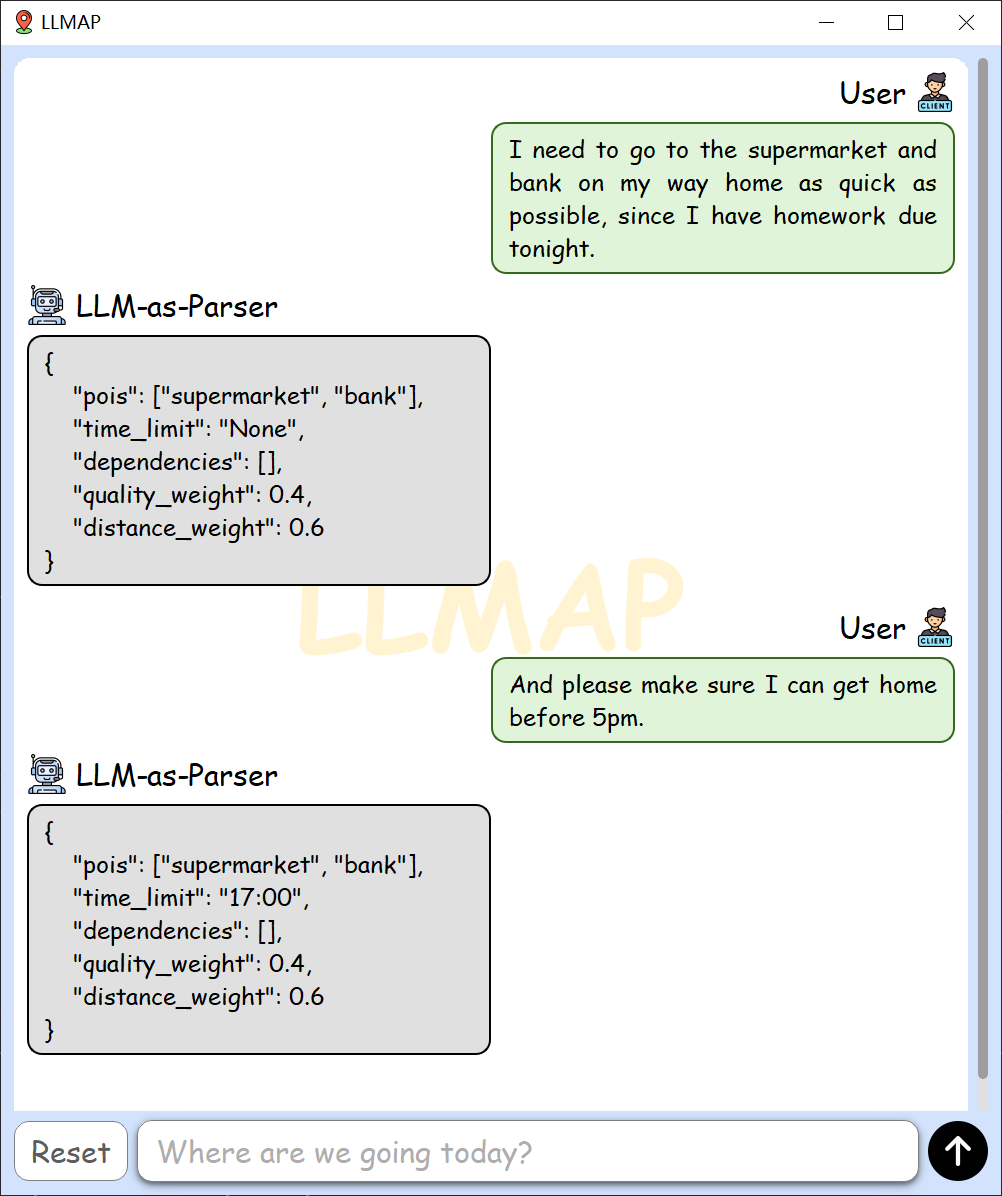}\label{fig:Correction 1}}
\hfill
\subfigure[Ambiguity]{\includegraphics[width=0.32\linewidth]{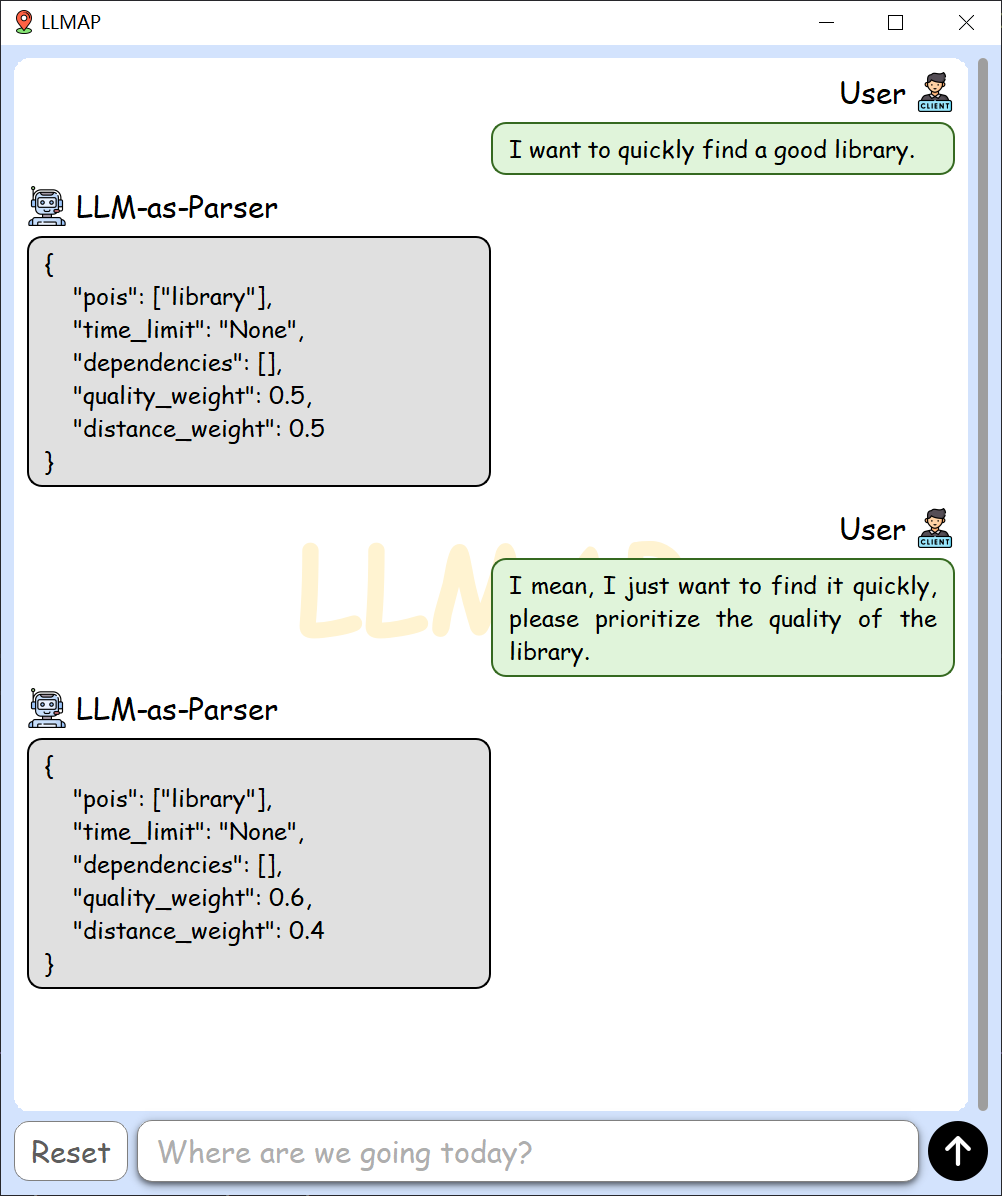}\label{fig:Correction 2}}
\hfill
\subfigure[Misinterpretation]{\includegraphics[width=0.32\linewidth]{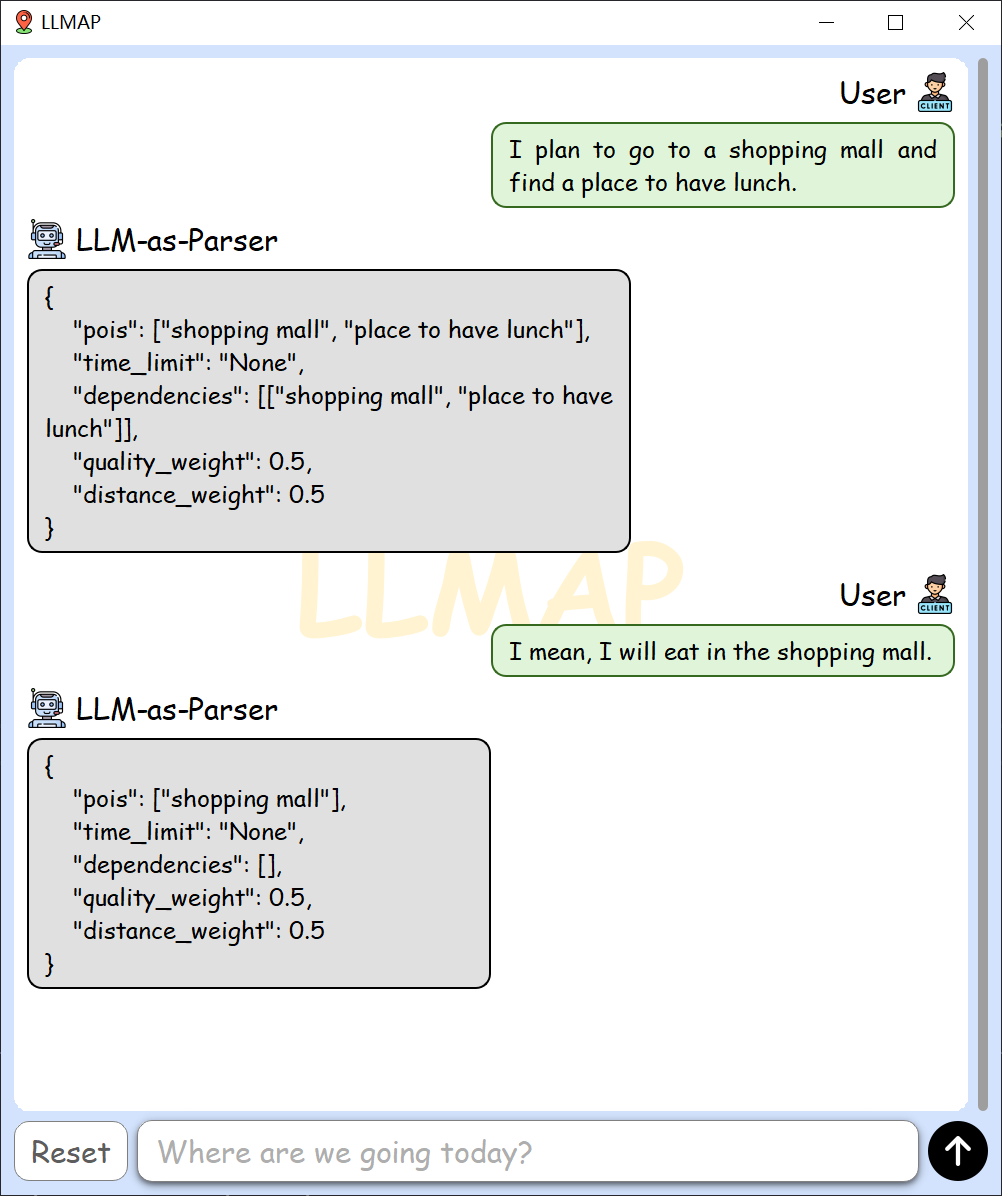}\label{fig:Correction 3}}
\vspace{-10pt}
\captionof{figure}{Demonstration of errors and conversational correction strategies. The interface facilitates interaction between real users and the LLM while maintaining contextual information to support conversational functionality.}
\label{fig:Error Correction}
\end{figure*}

\noindent\textbf{Impact of Language-to-Numerical Preference.} As illustrated in Figure \ref{fig:Impact_Weight}, the distribution of estimates follows similar patterns for both low user preferences (weights ranging from 0 to 0.3) and high user preferences (weights ranging from 0.7 to 1.0). Specifically, LLM estimations remain consistent regardless of the precise synthetic label values within these ranges. This pattern persists across different LLMs and remains consistent with or without CoT prompting. For balanced user preferences (weights ranging from 0.4 to 0.6), LLM estimates cluster around 0.5. While some models demonstrate more robust estimation capabilities (e.g., Gemma-2-9B, GPT-4o, OpenAI o1-mini, OpenAI o1) with shorter distribution tails, all LLMs exhibit a stepped distribution rather than the ideal linear trend (indicated by the gray dashed line). This reveals LLM ability to approximately classify user preferences, or more broadly, sentiments, from natural language, while struggling with fine-grained numerical outputs. This suggests that further fine-tuning might be necessary to better capture individual linguistic nuances. A sample and further analysis are provided in Appendix \ref{appendix:HIPP}.

\noindent\textbf{Impact of CoT on Parsing.} While CoT prompting significantly improves the performance of LLM-as-Parser for certain models (e.g., Phi-3-mini and Phi-3.5-mini), it adversely affects others (e.g., Gemma-2-2B (CoT)). The improvements in the former case can be attributed to the prevention of format errors that often result in default parameter usage without CoT prompting. However, in the latter case, CoT prompting leads to an overestimation of POI correlations. For example, given the instruction ``I want to go to the supermarket and the library today,'' there is no explicit dependency (visiting the supermarket before the library is not mandatory). Nevertheless, the sequential ordering in the sentence (``supermarket'' before ``library'') causes the model to overestimate their interdependence.

\subsection{User-to-LLM Conversational Correction}

We present user interaction scenarios in Figure \ref{fig:Error Correction}, demonstrating user engagement with the \Name system to identify and rectify computational reasoning errors through a graphical user interface implemented via the PyQt5 framework. Beyond the linguistic errors in Figures \ref{fig:Error 1}, \ref{fig:Error 2}, and \ref{fig:Error 3}, we present three representative LLM reasoning errors and corrective strategies. Figure \ref{fig:Correction 1} displays a user's vague instruction that fails to explicitly specify the exact time of the deadline, consequently hindering the LLM's ability to infer the time limit constraint. Figure \ref{fig:Correction 2} illustrates a case of ambiguous user instruction, where ``quickly'' modifies the action ``find'' rather than indicating that the route itself should be fast, resulting in a misinterpretation of user preference. In Figure \ref{fig:Correction 3}, the LLM misconstrues the user's intention of ``find a place'', where the phrase does not refer to a ``place to have lunch'' as a POI, but rather represents one purpose for visiting the shopping mall. These scenarios demonstrate that conversational correction effectively addresses LLM comprehension errors through iterative user dialogue.

\subsection{Further Analysis} 
We provide detailed analyses in Appendix \ref{appendix:Further Analysis}, including the time complexity analysis of \AlgName, empirical runtime comparisons, ablation study, transportation modes, departure days and times, the impact of geographic variations, and the integration of essential waypoints.

\section{Conclusion}

In this paper, we present a novel \Name system that integrates LLM-as-Parser with the \AlgName algorithm to facilitate multi-objective route planning. Our system decomposes the complex task into two components: using LLMs to understand and extract information from human instructions, and employing \AlgName to generate optimal routes that satisfy both constraints and user preferences. Through extensive experiments across 27 major cities in 14 countries, we demonstrate that our approach outperforms LLM-as-Agent solutions across multiple metrics. A future direction is to integrate richer information sources, such as user text reviews, as supplementary input to LLMs to enhance preference matching.

\section*{Limitations}

The primary limitation of our proposed method lies in the computational overhead of the \AlgName algorithm when handling a large number of POI types, as it requires permutation of POI type sets to find the optimal solution. Human instructions typically do not encompass an extensive list of POI types, as users rarely express intentions to visit dozens of locations such as hospitals, supermarkets, banks, bookstores, museums, and theaters in a single request. Nevertheless, the time complexity of the \AlgName algorithm does increase significantly with the number of POI types.

\section*{Acknowledgments}

This work was supported by a doctoral fellowship offered by the Purdue-Windracers Center AIDA\textsuperscript{3}, in part by the National Science Foundation (NSF) under grants CPS-2313109 and CPS-2146171, and by the OpenAI Researcher Access Program.

\bibliography{reference}

\onecolumn
\twocolumn
\setlength{\parindent}{1em}

\appendix

\section{Details of Experimental Setup}
\label{appendix:experimental setup}

\subsection{Default Settings}

LLMs exhibit potential reasoning errors, particularly in output formatting, which may precipitate system failures such as improper JSON format generation or missing dictionary keys that trigger subsequent errors. We therefore establish default formatting protocols to rectify any LLM output anomalies. Figure \ref{fig:Error} illustrates this phenomenon wherein Phi-3-mini fails to execute parsing behavior in accordance with user instructions during inference, instead generating hallucinated outputs containing inappropriate keys and semantically vacuous information. Under these circumstances, we implement error correction through default parametrization to ensure successful system execution. 

\begin{tcolorbox}[boxrule = 0.5pt, breakable, box align=center,]

\textbf{Original Output (Erroneous):}
\begin{lstlisting}[language=JSON,literate={
      {:}{{{\color{punct}{:}}}}{1}
      {,}{{{\color{punct}{,}}}}{1}
      {\{}{{{\color{RoyalBlue}{\{}}}}{1}
      {\}}{{{\color{RoyalBlue}{\}}}}}{1}
      {[}{{{\color{RoyalBlue}{[}}}}{1}
      {]}{{{\color{RoyalBlue}{]}}}}{1}}]
{
    "pois": ["poi1", "poi2", ...],
    "time_constraint": "HH:00",
    "dependencies": [["poi1", "poi2"], ...],
    "rating_weight": float,
    "route_weight": float,
    "weather_constraint": "weather_condition",
    "special_events": ["event1", "event2", ...],
    "traffic_condition": "light/moderate/heavy",
    "vehicle_type": "car/bike/public_transport"
}
\end{lstlisting}

\rule{\linewidth}{0.5pt}

\textbf{Error Correction (Default Setting):}
\begin{lstlisting}[language=JSON]
{
    "pois": [],
    "time_limit": "None",
    "dependencies": [],
    "quality_weight": 0.5,
    "distance_weight": 0.5
}
\end{lstlisting}
\end{tcolorbox}
\vspace{-10pt}

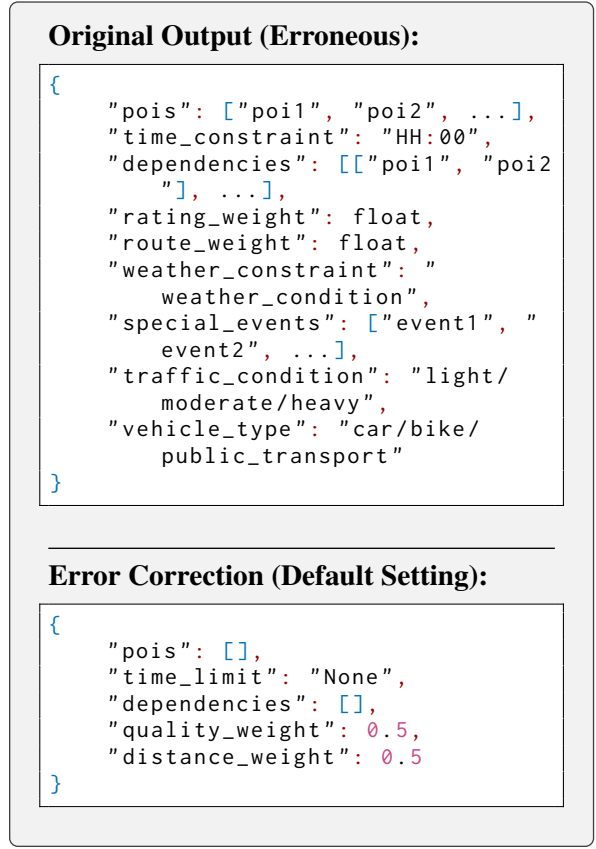
\captionof{figure}{An example of Phi-3-mini generating erroneous output. The model fails to adhere to expected schema, producing irrelevant keys and placeholders instead of proper values.}
\label{fig:Error}
\vspace{10pt}

\subsection{Time Parameters Setup}

We configure the default route departure time as Monday 10:00 AM in all experiments. The visit duration parameters are systematically assigned for different POI categories: 120 minutes for shopping malls, 30 minutes for supermarkets, 15 minutes for pharmacies, 20 minutes for banks, and 60 minutes for libraries. These duration values are incorporated into the total time calculation whenever the corresponding POIs are included in the generated routes.

This experimental configuration is designed to be flexible and extensible for parameter adjustment, serving primarily as a mechanism to evaluate route feasibility with respect to user time limits and POI operating hours. In practical applications, the system can be extended to incorporate real-time data by leveraging smartphone timestamps and geolocation services. Furthermore, the visit duration estimates can be refined by integrating Google Maps' real-time occupancy data for more accurate predictions based on venue congestion levels.

\subsection{Evaluation Metrics}
\label{appendix:Evaluation Metrics}
\textbf{Evaluation Metrics Formulation.} We evaluate the performance of our LLM-as-Parser approach across multiple dimensions: POI type identification, user time limit extraction, task dependency detection, and user preference weight estimation. For the set-based evaluations, we compute the F1 score by comparing the predictions against ground truth labels. Formally, for each instance $i$, we denote the ground truth set as $\mathbf{y}_i$ and the predicted set as $\hat{\mathbf{y}}_i$. The precision, recall, and F1 score are defined as:
\begin{equation}
\begin{aligned}
\text{Precision} &= \frac{1}{N}\sum_{i=1}^N\frac{|\hat{\mathbf{y}}_i \cap \mathbf{y}_i|}{|\hat{\mathbf{y}}_i|}, \\
\text{Recall} &= \frac{1}{N}\sum_{i=1}^N\frac{|\hat{\mathbf{y}}_i \cap \mathbf{y}_i|}{|\mathbf{y}_i|}, \\
F_1 &= \frac{2 \times \text{Precision} \times \text{Recall}}{\text{Precision} + \text{Recall}}, \\
\end{aligned}
\end{equation}
where $N$ denotes the number of samples. Furthermore, to assess the capability of our LLM-as-Parser in extracting user time limits and user preference weights, we employ two different evaluation metrics. For user time limit extraction, we define an accuracy metric. In this context, only two outcomes are possible: either the label and the estimated value are identical, yielding a score of 1, or they differ, yielding a score of 0. For each instance $i$, let $y_i$ denote the ground truth value and $\hat{y}_i$ the estimated value. This is formalized via the indicator function:
\begin{equation} \begin{aligned} \text{Accuracy} &= \frac{1}{N}\sum_{i=1}^{N} \mathbb{I}\{y_i = \hat{y}_i\}, \\ &\mathbb{I}\{y_i = \hat{y}_i\} = \begin{cases} 1, & \text{if } y_i = \hat{y}_i, \\ 0, & \text{otherwise}. \end{cases} \end{aligned} \end{equation}
For user preference weights, we define a similarity metric to measure how close the estimated weight distribution is to the ground truth. Let $\mathbf{w}_i = (w_i^1, w_i^2, \dots, w_i^d)$ denote the ground truth weight vector for instance $i$, and $\hat{\mathbf{w}}_i$ the estimated weight vector. The similarity is defined as:
\begin{equation}
\text{Similarity} = \frac{1}{N}\sum_{i=1}^{N} \left( 1 - \frac{1}{d} \sum_{j=1}^{d} \left\lvert w_i^j - \hat{w}_i^j \right\rvert \right),
\end{equation}
where $d$ is the number of weight components, indexed by $j$ (e.g., rating weight and route weight). This metric yields a value in $[0,1]$, with 1 indicating identical distributions and 0 indicating maximal difference.

\noindent\textbf{Table \ref{Table Color Setting} Visualization.} To facilitate better comparison of different methods across multiple objectives, we establish distinct thresholds for each metric as shown in Table \ref{Table Color Setting}, enhancing the visualization of results in Table \ref{Table Main Results}. These thresholds are determined based on empirical observations and result distribution, though they can be adjusted according to user requirements in different scenarios. For ratings, task completion rate, and user time limit, we implement multiple threshold levels to enable more nuanced distinctions. In contrast, we employ binary color coding for number of reviews, route length, dependency, and opening hours metrics. Through this color-coded visualization scheme, we aim to identify methods that satisfy all thresholds, indicated by blue coloring across all metrics.

\begin{table}[h]
\caption{Color threshold settings used in Table \ref{Table Main Results}. This color-coding scheme is purely for visualization purposes and does not affect any numerical results.}
\label{Table Color Setting}
\renewcommand{\arraystretch}{1.2}
\setlength{\aboverulesep}{0pt}
\setlength{\belowrulesep}{0pt}

\scriptsize
\begin{tabular}{cc}

\begin{minipage}[t]{0.2\textwidth}
\centering 
\begin{tabular}{!{\vrule width 0.25ex}>{\centering\arraybackslash}p{70pt}!{\vrule width 0.25ex}}
\toprule
\textbf{Rating} ($\uparrow$) \\
\toprule
\cellcolor{Yes} $\geq$ 4 \\
\cellcolor{No} 3.5 to 4 \\
\cellcolor{Noo} 3 to 3.5 \\
\cellcolor{Nooo} $\leq$ 3 \\
\toprule
\end{tabular}
\end{minipage}
&
\begin{minipage}[t]{0.2\textwidth}
\centering 
\begin{tabular}{!{\vrule width 0.25ex}>{\centering\arraybackslash}p{70pt}!{\vrule width 0.25ex}}
\toprule
\textbf{Number of Review} ($\uparrow$) \\
\toprule
\cellcolor{Yes} $\geq$ 1000 \\
\cellcolor{No} $\leq$ 1000 \\
\toprule
\end{tabular}
\end{minipage}
\\[40pt]

\begin{minipage}[t]{0.2\textwidth}
\centering 
\begin{tabular}{!{\vrule width 0.25ex}>{\centering\arraybackslash}p{70pt}!{\vrule width 0.25ex}}
\toprule
\textbf{Task Completion} \\
\textbf{Rate} ($\uparrow$) \\
\toprule
\cellcolor{Yes} $\geq$ 90\% \\
\cellcolor{No} 80\% to 90\% \\
\cellcolor{Noo} 70\% to 80\% \\
\cellcolor{Nooo} $\leq$ 70\% \\
\toprule
\end{tabular}
\end{minipage}
&
\begin{minipage}[t]{0.2\textwidth}
\centering 
\begin{tabular}{!{\vrule width 0.25ex}>{\centering\arraybackslash}p{70pt}!{\vrule width 0.25ex}}
\toprule
\textbf{Route Length} ($\downarrow$) \\
\toprule
\cellcolor{Yes} $\leq$ 30 km \\
\cellcolor{No} $\geq$ 30 km \\
\toprule
\end{tabular}
\end{minipage}
\\[50pt]

\begin{minipage}[t]{0.2\textwidth}
\centering 
\begin{tabular}{!{\vrule width 0.25ex}>{\centering\arraybackslash}p{70pt}!{\vrule width 0.25ex}}
\toprule
\textbf{Time Limit} ($\downarrow$) \\
\toprule
\cellcolor{Yes} $\leq$ 0 hour \\
\cellcolor{No} $0$ to $\sfrac{1}{3}$ hours \\
\cellcolor{Noo} $\sfrac{1}{3}$ to $1$ hour \\
\cellcolor{Nooo} $\geq$ 1 hour \\
\toprule
\end{tabular}
\end{minipage}
&
\begin{minipage}[t]{0.2\textwidth}
\centering 
\begin{tabular}{!{\vrule width 0.25ex}>{\centering\arraybackslash}p{70pt}!{\vrule width 0.25ex}}
\toprule
\textbf{Dependency} ($\downarrow$) \\
\toprule
\cellcolor{Yes} $\leq$ 0\% \\
\cellcolor{No} $\geq$ 0\% \\
\toprule
\end{tabular}
\end{minipage}
\\[50pt]

&
\begin{minipage}[t]{0.2\textwidth}
\centering 
\begin{tabular}{!{\vrule width 0.25ex}>{\centering\arraybackslash}p{70pt}!{\vrule width 0.25ex}}
\toprule
\textbf{Opening Hours} ($\downarrow$) \\
\toprule
\cellcolor{Yes} $\leq$ 0\% \\
\cellcolor{No} $\geq$ 0\% \\
\toprule
\end{tabular}
\end{minipage}
\\

\end{tabular}
\end{table}

\newpage

\section{Details of Dataset}
\label{appendix:dataset}

\subsection{HIPP Dataset}
\label{appendix:HIPP}

The HIPP dataset comprises 1,000 evaluation samples, each containing a unique synthetic label, human instruction, and 24 LLM-as-Parser estimations. Figure \ref{fig:HIPP} illustrates a representative example from the dataset. Our analysis reveals that most LLMs successfully perform their basic functions (i.e., extracting POI types and constraints), with the exception of Phi-3-mini and Phi-3.5-mini. These exceptions stem from their inability to generate information in the correct format (i.e., JSON format). To ensure that route planning does not fail, we implement default fallback settings for these models, as demonstrated in Figure \ref{fig:HIPP}. Under these fallback settings, the route is simplified to direct travel between the origin and destination without any intermediate POI visits. We observe that CoT prompting enhances the ability of Phi-3-mini and Phi-3.5-mini to output correctly formatted responses. While most models accurately extract POI types, user time limits, and dependency constraints, they exhibit uncertainty in identifying user preference weights, as illustrated in Figure \ref{fig:Impact_Weight}.

\onecolumn

\begin{tcolorbox}[boxrule = 0.5pt, breakable, box align=center,]
\textbf{Synthetic Label:}

\begin{lstlisting}[language=JSON]
{
    "pois": ["bank", "library", "supermarket", "shopping mall"],
    "time_limit": "19:00",
    "dependencies": [["bank", "library"], ["library", "supermarket"]],
    "quality_weight": 0.9,
    "distance_weight": 0.1
}
\end{lstlisting}

\rule{\linewidth}{0.5pt}

\textbf{Human Instruction (Synthetic Data):}

Today, let's plan to visit the bank, library, supermarket, and shopping mall. Please be home by 19:00. Prioritize visiting POIs with high ratings as they are more important today. Start at the bank before heading to the library, and continue to the supermarket right after the library.

\rule{\linewidth}{0.5pt}

\textbf{LLM-as-Parser (Estimation):}

\begin{lstlisting}[language=JSON,escapeinside={(*}{*)}]
(*\textcolor{key}{"Phi-3-mini"}*):
{
    "pois": [],
    "time_limit": "None",
    "dependencies": [],
    "quality_weight": 0.5,
    "distance_weight": 0.5
}
\end{lstlisting}

\begin{lstlisting}[language=JSON,escapeinside={(*}{*)}]
(*\textcolor{key}{"Phi-3.5-mini"}*):
{
    "pois": [],
    "time_limit": "None",
    "dependencies": [],
    "quality_weight": 0.5,
    "distance_weight": 0.5
}
\end{lstlisting}

\begin{lstlisting}[language=JSON,escapeinside={(*}{*)}]
(*\textcolor{key}{"LLaMA-3.2-3B"}*):
{
    "pois": ["bank", "library", "supermarket", "shopping mall"],
    "time_limit": "19:00",
    "dependencies": [["bank", "library"], ["library", "supermarket"]],
    "quality_weight": 0.7,
    "distance_weight": 0.3
}
\end{lstlisting}

\begin{lstlisting}[language=JSON,escapeinside={(*}{*)}]
(*\textcolor{key}{"LLaMA-3.1-8B"}*):
{
   "pois": ["bank", "library", "supermarket", "shopping mall"],
   "time_limit": "19:00",
   "dependencies": [["bank", "library"], ["library", "supermarket"]],
   "quality_weight": 0.7,
   "distance_weight": 0.3
}
\end{lstlisting}

\begin{lstlisting}[language=JSON,escapeinside={(*}{*)}]
(*\textcolor{key}{"Mistral-7B-v0.3"}*):
{
   "pois": ["bank", "library", "supermarket", "shopping mall"],
   "time_limit": "19:00",
   "dependencies": [["bank", "library"], ["library", "supermarket"]],
   "quality_weight": 0.7,
   "distance_weight": 0.3
}
\end{lstlisting}

\begin{lstlisting}[language=JSON,escapeinside={(*}{*)}]
(*\textcolor{key}{"Gemma-2-2B"}*):
{
   "pois": ["bank", "library", "supermarket", "shopping mall"],
   "time_limit": "19:00",
   "dependencies": [["bank", "library"], ["library", "supermarket"]],
   "quality_weight": 0.8,
   "distance_weight": 0.2
}
\end{lstlisting}

\begin{lstlisting}[language=JSON,escapeinside={(*}{*)}]
(*\textcolor{key}{"Gemma-2-9B"}*):
{
   "pois": ["bank", "library", "supermarket", "shopping mall"],
   "time_limit": "19:00",
   "dependencies": [["bank", "library"], ["library", "supermarket"]],
   "quality_weight": 0.8,
   "distance_weight": 0.2
}
\end{lstlisting}

\begin{lstlisting}[language=JSON,escapeinside={(*}{*)}]
(*\textcolor{key}{"GPT-3.5"}*):
{
   "pois": ["bank", "library", "supermarket", "shopping mall"],
   "time_limit": "19:00",
   "dependencies": [["bank", "library"], ["library", "supermarket"]],
   "quality_weight": 0.7,
   "distance_weight": 0.3
}
\end{lstlisting}

\begin{lstlisting}[language=JSON,escapeinside={(*}{*)}]
(*\textcolor{key}{"GPT-4o-mini"}*):
{
   "pois": ["bank", "library", "supermarket", "shopping mall"],
   "time_limit": "19:00",
   "dependencies": [["bank", "library"], ["library", "supermarket"]],
   "quality_weight": 0.75,
   "distance_weight": 0.25
}
\end{lstlisting}

\begin{lstlisting}[language=JSON,escapeinside={(*}{*)}]
(*\textcolor{key}{"GPT-4o"}*):
{
   "pois": ["bank", "library", "supermarket", "shopping mall"],
   "time_limit": "19:00",
   "dependencies": [["bank", "library"], ["library", "supermarket"]],
   "quality_weight": 0.7,
   "distance_weight": 0.3
}
\end{lstlisting}

\begin{lstlisting}[language=JSON,escapeinside={(*}{*)}]
(*\textcolor{key}{"OpenAI o1-mini"}*):
{
   "pois": ["bank", "library", "supermarket", "shopping mall"],
   "time_limit": "19:00",
   "dependencies": [["bank", "library"], ["library", "supermarket"]],
   "quality_weight": 0.7,
   "distance_weight": 0.3
}
\end{lstlisting}

\begin{lstlisting}[language=JSON,escapeinside={(*}{*)}]
(*\textcolor{key}{"OpenAI o1"}*):
{
   "pois": ["bank", "library", "supermarket", "shopping mall"],
   "time_limit": "19:00",
   "dependencies": [["bank", "library"], ["library", "supermarket"]],
   "quality_weight": 0.7,
   "distance_weight": 0.3
}
\end{lstlisting}

\begin{lstlisting}[language=JSON,escapeinside={(*}{*)}]
(*\textcolor{key}{"Phi-3-mini (CoT)"}*):
{
   "pois": ["bank", "library", "supermarket", "shopping mall"],
   "time_limit": "19:00",
   "dependencies": [["bank", "library"], ["library", "supermarket"]],
   "quality_weight": 0.7,
   "distance_weight": 0.3
}
\end{lstlisting}

\begin{lstlisting}[language=JSON,escapeinside={(*}{*)}]
(*\textcolor{key}{"Phi-3.5-mini (CoT)"}*):
{
   "pois": ["bank", "library", "supermarket", "shopping mall"],
   "time_limit": "19:00",
   "dependencies": [["bank", "library"], ["library", "supermarket"]],
   "quality_weight": 0.5,
   "distance_weight": 0.5
}
\end{lstlisting}

\begin{lstlisting}[language=JSON,escapeinside={(*}{*)}]
(*\textcolor{key}{"LLaMA-3.2-3B (CoT)"}*):
{
   "pois": ["bank", "library", "supermarket", "shopping mall"],
   "time_limit": "19:00",
   "dependencies": [["bank", "library"], ["library", "supermarket"]],
   "quality_weight": 1.0,
   "distance_weight": 1.0
}
\end{lstlisting}

\begin{lstlisting}[language=JSON,escapeinside={(*}{*)}]
(*\textcolor{key}{"LLaMA-3.1-8B (CoT)"}*):
{
   "pois": ["bank", "library", "supermarket", "shopping mall"],
   "time_limit": "19:00",
   "dependencies": [["bank", "library"], ["library", "supermarket"]],
   "quality_weight": 1.0,
   "distance_weight": 0.0
}
\end{lstlisting}

\begin{lstlisting}[language=JSON,escapeinside={(*}{*)}]
(*\textcolor{key}{"Mistral-7B-v0.3 (CoT)"}*):
{
   "pois": ["bank", "library", "supermarket", "shopping mall"],
   "time_limit": "19:00",
   "dependencies": [["bank", "library"], ["library", "supermarket"]],
   "quality_weight": 0.7,
   "distance_weight": 0.3
}
\end{lstlisting}

\begin{lstlisting}[language=JSON,escapeinside={(*}{*)}]
(*\textcolor{key}{"Gemma-2-2B (CoT)"}*):
{
   "pois": ["bank", "library", "supermarket", "shopping mall"],
   "time_limit": "19:00",
   "dependencies": [["bank", "library"], ["library", "supermarket"]],
   "quality_weight": 0.7,
   "distance_weight": 0.3
}
\end{lstlisting}

\begin{lstlisting}[language=JSON,escapeinside={(*}{*)}]
(*\textcolor{key}{"Gemma-2-9B (CoT)"}*):
{
   "pois": ["bank", "library", "supermarket", "shopping mall"],
   "time_limit": "19:00",
   "dependencies": [["bank", "library"], ["library", "supermarket"]],
   "quality_weight": 0.7,
   "distance_weight": 0.3
}
\end{lstlisting}

\begin{lstlisting}[language=JSON,escapeinside={(*}{*)}]
(*\textcolor{key}{"GPT-3.5 (CoT)"}*):
{
   "pois": ["bank", "library", "supermarket", "shopping mall"],
   "time_limit": "19:00",
   "dependencies": [["bank", "library"], ["library", "supermarket"]],
   "quality_weight": 0.7,
   "distance_weight": 0.3
}
\end{lstlisting}

\begin{lstlisting}[language=JSON,escapeinside={(*}{*)}]
(*\textcolor{key}{"GPT-4o (CoT)"}*):
{
   "pois": ["bank", "library", "supermarket", "shopping mall"],
   "time_limit": "19:00",
   "dependencies": [["bank", "library"], ["library", "supermarket"]],
   "quality_weight": 0.7,
   "distance_weight": 0.3
}
\end{lstlisting}

\begin{lstlisting}[language=JSON,escapeinside={(*}{*)}]
(*\textcolor{key}{"GPT-4o-mini (CoT)"}*):
{
   "pois": ["bank", "library", "supermarket", "shopping mall"],
   "time_limit": "19:00",
   "dependencies": [["bank", "library"], ["library", "supermarket"]],
   "quality_weight": 0.7,
   "distance_weight": 0.3
}
\end{lstlisting}

\begin{lstlisting}[language=JSON,escapeinside={(*}{*)}]
(*\textcolor{key}{"OpenAI o1-mini (CoT)"}*):
{
   "pois": ["bank", "library", "supermarket", "shopping mall"],
   "time_limit": "19:00",
   "dependencies": [["bank", "library"], ["library", "supermarket"]],
   "quality_weight": 0.7,
   "distance_weight": 0.3
}
\end{lstlisting}

\begin{lstlisting}[language=JSON,escapeinside={(*}{*)}]
(*\textcolor{key}{"OpenAI o1 (CoT)"}*):
{
   "pois": ["bank", "library", "supermarket", "shopping mall"],
   "time_limit": "19:00",
   "dependencies": [["bank", "library"], ["library", "supermarket"]],
   "quality_weight": 0.7,
   "distance_weight": 0.3
}
\end{lstlisting}
\end{tcolorbox}
\vspace{-10pt}
\captionof{figure}{A representative example illustrating the relationship between synthetic labels, human instructions, and LLM-as-Parser estimations.}
\label{fig:HIPP}

\twocolumn
\setlength{\parindent}{1em}

\subsection{Map Service and Scenario Setting}
\label{appendix:map service}

\textbf{Map Service.} For POI data acquisition, we utilize the Google Places API to retrieve place (i.e., POI) information. The API accepts coordinates as input and returns comprehensive POI data within a specified search radius for given place types (i.e., POI types). The returned information includes POI IDs, names, ratings, numbers of ratings, geographical coordinates, etc. We set the end point as the search center with a radius of 5000 meters. Since the Google Places API limits each query (referred to as a "page") to 20 POI entries, we iterate through all available pages to collect complete POI information for the specified area. Notably, our system is not limited to any specific map service API, it can be readily adapted to work with various mapping services (e.g., aviation maps, hiking maps) to support diverse route planning tasks.

\noindent\textbf{Incomplete POI Data.} The real-world data used in this paper, sourced from Google Places API, while not producing formatically inaccurate data (e.g., ratings outside the 1-5 range) and typically possessing geographical location data, presents instances of incomplete information, particularly in regions where Google Maps experiences lower popularity. Consequently, we implement default configuration settings to mitigate system operational failures attributable to incomplete data. Specifically, we configure any incomplete data as follows: ratings as 1.0 (i.e., the minimum value), number of reviews as 1 (i.e., the minimum value), and opening hours as standard business hours (9:00 AM to 5:00 PM). An extreme case occurs when all POIs within a region lack any data, resulting in route planning that relies entirely on distance without considering other user preferences. These configurations aim to minimize potential risks in route planning, and researchers can readily modify these default settings based on empirical data.

\noindent\textbf{Use Case Scenario.} To demonstrate our \Name system's location-independent applicability and global generalizability, we conduct extensive testing across 27 major cities in 14 countries worldwide, as shown in Table \ref{Table Scenario}. We focus on a common use scenario: students returning to university from airports with intermediate POI stops for various activities. This scenario serves as a representative example that can be readily extended to other situations, such as airport-to-hotel transfers, commutes from office to home, or travel between home and school. Furthermore, our proposed algorithm can be effectively deployed in scenarios where the start and end points share the same location, such as round trips from home to grocery stores. Such cases are handled by treating identical locations as distinct nodes with the same coordinates.

\noindent\textbf{Extended Transportation Scenarios.} Our system's versatility extends beyond specific transportation modes, encompassing various mobility scenarios including walking (e.g., city walk, hiking trails), public transit (e.g., urban transit systems, aviation networks), and logistics delivery. A key advantage of our approach is its training-free nature, requiring only parameter adjustments to adapt to different use cases. For example, consider travel planning where users need to coordinate flights between multiple destination cities. In this context, the system can optimize multiple objectives such as minimizing flight duration and costs while maximizing flight quality (e.g., avoiding budget airlines), subject to constraints including total trip duration, city visit sequence, and flight availability. Given access to flight service data, our \Name system and \AlgName algorithm can reliably generate optimal routes while satisfying all specified constraints.

\onecolumn

\begin{sidewaystable}[h!]
\centering
\resizebox{0.9\linewidth}{!}{
\begin{tabular}{ll|lcc|lcc}
\toprule
\multirow{2}{*}{\textbf{Country}} & \multirow{2}{*}{\textbf{City}} & \multicolumn{3}{c|}{\textbf{Start Point}} & \multicolumn{3}{c}{\textbf{End Point}} \\
& & Name & Latitude & Longitude & Name & Latitude & Longitude \\
\midrule

USA & New York & John F. Kennedy International Airport & 40.644624 & -73.779703 & Columbia University & 40.807536 & -73.962573 \\
USA & New York & John F. Kennedy International Airport & 40.644624 & -73.779703 & New York University & 40.712775 & -74.005973 \\
USA & Los Angeles & Los Angeles International Airport & 33.942153 & -118.403605 & University of California, Los Angeles & 34.069918 & -118.443849 \\
USA & Chicago & O'Hare International Airport & 41.980259 & -87.908986 & University of Chicago & 41.790448 & -87.600395 \\
USA & Chicago & O'Hare International Airport & 41.980259 & -87.908986 & Northwestern University & 42.056459 & -87.675267 \\
USA & San Francisco & San Francisco International Airport & 37.619115 & -122.381627 & University of California, Berkeley & 37.871214 & -122.255463 \\
USA & San Jose & San Jose International Airport & 37.363529 & -121.928593 & Stanford University & 37.427660 & -122.170060 \\
USA & Seattle & Seattle-Tacoma International Airport & 47.448365 & -122.308593 & University of Washington & 47.656717 & -122.306618 \\
USA & Boston & Logan International Airport & 42.365602 & -71.009614 & Harvard University & 42.374437 & -71.118249 \\
USA & Boston & Logan International Airport & 42.365602 & -71.009614 & Massachusetts Institute of Technology & 42.360091 & -71.094160 \\
Canada & Toronto & Toronto Pearson International Airport & 43.679834 & -79.628383 & University of Toronto & 43.661541 & -79.400875 \\
Canada & Vancouver & Vancouver International Airport & 49.193374 & -123.175128 & University of British Columbia & 49.260605 & -123.245994 \\
China & Beijing & Beijing Capital International Airport & 40.079857 & 116.603112 & Peking University & 39.986913 & 116.305874 \\
China & Beijing & Beijing Capital International Airport & 40.079857 & 116.603112 & Tsinghua University & 40.006158 & 116.318407 \\
China & Shanghai & Shanghai Pudong International Airport & 31.144344 & 121.808273 & Fudan University & 31.297420 & 121.503618 \\
China & Shanghai & Shanghai Pudong International Airport & 31.144344 & 121.808273 & Shanghai Jiao Tong University & 31.025220 & 121.433778 \\
China & Hong Kong & Hong Kong International Airport & 22.313474 & 113.913728 & University of Hong Kong & 22.283089 & 114.136562 \\
China & Hong Kong & Hong Kong International Airport & 22.313474 & 113.913728 & Chinese University of Hong Kong & 22.419625 & 114.206761 \\
Taiwan & Taipei & Taiwan Taoyuan International Airport & 25.080490 & 121.231159 & National Taiwan University & 25.017340 & 121.539752 \\
Japan & Tokyo & Tokyo Narita International Airport & 35.770178 & 140.384321 & University of Tokyo & 35.713816 & 139.762734 \\
South Korea & Seoul & Incheon International Airport & 37.458666 & 126.441968 & Seoul National University & 37.464827 & 126.957199 \\
South Korea & Seoul & Incheon International Airport & 37.458666 & 126.441968 & Yonsei University & 37.566394 & 126.938707 \\
Singapore & Singapore & Changi Airport & 1.358604 & 103.989944 & National University of Singapore & 1.296643 & 103.776394 \\
Singapore & Singapore & Changi Airport & 1.358604 & 103.989944 & Nanyang Technological University & 1.348310 & 103.683135 \\
India & Mumbai & Chhatrapati Shivaji Maharaj International Airport & 19.090218 & 72.862812 & Indian Institute of Technology Bombay & 19.133060 & 72.915106 \\
India & Delhi & Indira Gandhi International Airport & 28.556144 & 77.099962 & Indian Institute of Technology Delhi & 28.545718 & 77.192768 \\
Australia & Sydney & Sydney Kingsford Smith Airport & -33.950031 & 151.181694 & University of Sydney & -33.888404 & 151.186765 \\
Australia & Melbourne & Melbourne Tullamarine Airport & -37.669919 & 144.840345 & University of Melbourne & -37.798346 & 144.960974 \\
UK & London & London Heathrow Airport & 51.467991 & -0.455051 & Imperial College London & 51.498822 & -0.174873 \\
UK & London & London Heathrow Airport & 51.467991 & -0.455051 & University College London & 51.524559 & -0.134040 \\
UK & Oxford & London Heathrow Airport & 51.467991 & -0.455051 & University of Oxford & 51.757043 & -1.254518 \\
UK & Cambridge & London Heathrow Airport & 51.467991 & -0.455051 & University of Cambridge & 52.205356 & 0.113168 \\
France & Paris & Charles de Gaulle Airport & 49.007883 & 2.550785 & Sorbonne University & 48.846950 & 2.355570 \\
France & Paris & Charles de Gaulle Airport & 49.007883 & 2.550785 & Ecole Normale Superieure & 48.842024 & 2.344430 \\
Germany & Munich & Munich International Airport & 48.353987 & 11.788362 & Technical University of Munich & 48.148765 & 11.568176 \\
Italy & Milan & Milan Malpensa Airport & 45.622714 & 8.728234 & Politecnico di Milano & 45.468503 & 9.182403 \\
Russia & Moscow & Sheremetyevo International Airport & 55.973648 & 37.412503 & Moscow State University & 55.703935 & 37.528669 \\
\bottomrule
\end{tabular}
}
\caption{Start points and end points information across 27 major cities in 14 countries.}
\label{Table Scenario}
\end{sidewaystable}

\twocolumn
\setlength{\parindent}{1em}

\section{Further Analysis}
\label{appendix:Further Analysis}

\subsection{Time Complexity Analysis}
\label{appendix:Complexity}

Following Algorithm \ref{alg:MSGS}, we analyze the time complexity of our \AlgName algorithm in detail. Let $k = |\mathcal{Y}|$ represent the number of node types, with each node type containing the average $m = \frac{|\mathcal{V}|}{|\mathcal{Y}|}$ nodes. First, enumerating all possible node type orders introduces up to $k!$ permutations. In addition, exploring subsets of these $k$ node types involves up to $2^k$ subsets, each potentially requiring $(k-r)!$ permutations for further exploration. This results in an overall exponential complexity driven by $k$. Second, for each node type order, the graph construction has a polynomial time complexity $\mathcal{O}(km^2)$, while the Dijkstra search operates with a complexity of $\mathcal{O}\left((km)^2 \log(km)\right)$. Since $(km)^2 \log(km)$ dominates $km^2$, the worst-case complexity can be simplified to $\mathcal{O}\left(2^k \cdot k! \cdot (km)^2 \log(km)\right)$.

Considering the early stop mechanism (Algorithm \ref{alg:MSGS}, Line 10), the algorithm progressively explores combinations from $k$ node types to $1$ node type until finding a valid solution. The best-case time complexity becomes $\mathcal{O}\left(k! \cdot (km)^2 \log(km)\right)$ when a valid route containing all POI types is found immediately. The worst-case time complexity remains $\mathcal{O}\left(2^k \cdot k! \cdot (km)^2 \log(km)\right)$ if the algorithm needs to explore all possible combinations. In our routing scenario, the best case occurs when the planned route satisfies all constraints while including all POI types, while the worst case happens when including any POI type violates the constraints. According to Table \ref{Table Main Results}, more than 95\% of the cases successfully include all requested POI types, indicating that our algorithm typically achieves near best-case performance in practice.

\subsection{Runtime Analysis}
\label{appendix:Runtime}

Beyond the theoretical time complexity analysis, our empirical runtime measurements in Table \ref{tab:runtime_analysis} demonstrate that the \Name system significantly outperforms baseline approaches in computational efficiency. In contrast to LLM-as-Agent, our method utilizes LLMs solely for parsing human instructions, effectively limiting input length and reducing execution time. Although the computational cost of LLM-as-Agent scales linearly with the number of POIs, our approach remains independent of POI counts. Solver-based methods struggle with our optimization problem due to its multi-step, multi-objective, and multi-constraint nature, particularly given the conflicting objectives. To address these challenges, \Name employs a modular approach where different components serve distinct purposes: LLMs parse human instructions, while \AlgName constructs subgraphs incrementally to satisfy multiple constraints and objectives. Moreover, we optimize the time complexity of \AlgName through several key tricks, including early dependency constraint checking, omitting edges between nodes of the same type, and early stop criteria.

\begin{table}[t]
\centering
\scriptsize
\resizebox{\linewidth}{!}{
\begin{tabular}{l|ccc}
\toprule
\multirow{2.1}{*}{\textbf{Method}} & \textbf{Parsing} & \textbf{Planning} & \textbf{Total} \\
& \textbf{Time} (s) & \textbf{Time} (s) & \textbf{Time} (s) \\
\midrule
LLM-as-Agent & - & 18.89 & 18.89 \\
SMT Solver & 1.49 & 9.32 & 10.81 \\
SMT Solver v2 & 1.49 & 28.38 & 29.87 \\
\rowcolor{Yes} Ours & 1.49 & 0.29 & 1.78 \\
\bottomrule 
\end{tabular}
}
\caption{Runtime analysis of \Name and baseline approaches on LLaMA-3.2-3B (CoT). Parsing time refers to the runtime of LLM-as-Parser, while planning time denotes the runtime of our \AlgName algorithm or SMT solver execution.}
\label{tab:runtime_analysis}
\end{table}

\begin{table*}[t]
\centering

\resizebox{\linewidth}{!}{
\begin{tabular}{l|cccc|ccc}
\toprule
\textbf{Average Speed} & \textbf{Rating} & \textbf{Number of} & \textbf{Length} & \textbf{Task Completion} & \multicolumn{3}{c}{\textbf{Constraint Violation} ($\downarrow$)} \\
(km/h) & ($\uparrow$) & \textbf{Review} ($\uparrow$) & (km) ($\downarrow$) & \textbf{Rate} (\%) ($\uparrow$) & \textbf{Time Limit} (hrs) & \textbf{Dependency} (\%) &  \textbf{Opening Hours} (\%) \\
\midrule
5 (walking) & 3.29 & 5178 & 28.97 & 61.11 & 0.00 & 0.00 & 0.00 \\
10 (bicycle) & 4.06 & 5653 & 29.54 & 88.09 & 0.00 & 0.00 & 0.00 \\
20 (bus) & 4.28 & 5676 & 29.72 & 95.83 & 0.00 & 0.00 & 0.00 \\
30 (car) & 4.28 & 5530 & 29.73 & 96.44 & 0.00 & 0.00 & 0.00 \\
40 & 4.26 & 5133 & 29.71 & 95.99 & 0.00 & 0.00 & 0.00 \\
50 (UAV) & 4.25 & 5072 & 29.71 & 95.81 & 0.00 & 0.00 & 0.00 \\
60 & 4.25 & 5052 & 29.69 & 95.77 & 0.00 & 0.00 & 0.00 \\
\bottomrule
\end{tabular}
}
\caption{Impact of different transportation modes.}
\label{Table Transportation}

\vspace{0.8em}

\resizebox{\linewidth}{!}{
\begin{tabular}{l|cccc|ccc}
\toprule
\multirow{2.1}{*}{\textbf{Departure Day}} & \textbf{Rating} & \textbf{Number of} & \textbf{Length} & \textbf{Task Completion} & \multicolumn{3}{c}{\textbf{Constraint Violation} ($\downarrow$)} \\
& ($\uparrow$) & \textbf{Review} ($\uparrow$) & (km) ($\downarrow$) & \textbf{Rate} (\%) ($\uparrow$) & \textbf{Time Limit} (hrs) & \textbf{Dependency} (\%) &  \textbf{Opening Hours} (\%) \\
\midrule
Monday & 4.28 & 5530 & 29.73 & 96.44 & 0.00 & 0.00 & 0.00 \\
Tuesday & 4.28 & 4719 & 29.72 & 96.51 & 0.00 & 0.00 & 0.00 \\
Saturday & 4.00 & 5711 & 29.31 & 75.84 & 0.00 & 0.00 & 0.00 \\
Sunday & 3.41 & 7273 & 28.93 & 50.59 & 0.00 & 0.00 & 0.00 \\
\bottomrule
\end{tabular}
}
\caption{Impact of different departure days.}
\label{Table Departure Day}

\vspace{0.8em}

\resizebox{\linewidth}{!}{
\begin{tabular}{l|cccc|ccc}
\toprule
\textbf{Departure Time} & \textbf{Rating} & \textbf{Number of} & \textbf{Length} & \textbf{Task Completion} & \multicolumn{3}{c}{\textbf{Constraint Violation} ($\downarrow$)} \\
(HH:MM) & ($\uparrow$) & \textbf{Review} ($\uparrow$) & (km) ($\downarrow$) & \textbf{Rate} (\%) ($\uparrow$) & \textbf{Time Limit} (hrs) & \textbf{Dependency} (\%) &  \textbf{Opening Hours} (\%) \\
\midrule
08:00 & 3.72 & 2039 & 29.29 & 75.44 & 0.00 & 0.00 & 0.00 \\
10:00 & 4.28 & 5530 & 29.73 & 96.44 & 0.00 & 0.00 & 0.00 \\
12:00 & 4.28 & 5852 & 29.70 & 95.53 & 0.00 & 0.00 & 0.00 \\
14:00 & 4.18 & 5502 & 29.50 & 88.76 & 0.00 & 0.00 & 0.00 \\
16:00 & 3.68 & 5521 & 29.11 & 64.17 & 0.00 & 0.00 & 0.00 \\
18:00 & 3.06 & 5987 & 28.74 & 43.11 & 0.00 & 0.00 & 0.00 \\
\bottomrule
\end{tabular}
}
\caption{Impact of different departure times.}
\label{Table Departure Time}
\end{table*}

\subsection{Ablation Study}

We present an ablation study in Table \ref{Table Main Results} to evaluate the performance impact of removing two key components from our \Name system: LLM-as-Parser and \AlgName. For the former, we replace LLM-as-Parser outputs with synthetic labels, while for the latter, we directly apply the Dijkstra algorithm on the complete graph without considering the multi-step subgraph construction process. The synthetic labels can be viewed as an approximate ideal parsing scenario, though they do not represent a performance upper bound since user preference weights influence route evaluation metrics. We observe that \Name achieves comparable performance to this ideal scenario across multiple LLMs, as highlighted in Table \ref{Table Main Results} (e.g., GPT-4o, Phi-3-mini (CoT), Phi-3.5-mini (CoT), Mistral-7B (CoT), Gemma-2-2B (CoT), and GPT-4o (CoT)), validating the effectiveness of our LLM-as-Parser approach.

Regarding the w/o \AlgName ablation, this variant fails to generate meaningful routes, instead producing direct paths from the start point to the end point. This occurs because applying the Dijkstra algorithm directly on the graph focuses solely on minimizing weighted distances (based on ratings, number of reviews, and length) without maximizing task completion rates or considering constraints. This demonstrates the necessity of \AlgName's multi-step multi-objective optimization approach, which first maximizes task completion rates under constraints before trade-off multiple objectives.

\subsection{Impact of Transportation Mode, Departure Day, and Departure Time}

Our \Name system demonstrates substantial flexibility in accommodating diverse transportation configurations, analogous to Google Maps functionality, thereby enabling users to dynamically calibrate parameters in accordance with their specific requirements. We conduct comprehensive experiments to examine the influence of transportation modalities, departure scheduling, and temporal constraints on system performance metrics.

First, transportation modalities characterized by varying average velocities exhibit a pronounced correlation between mobility speed and task completion efficacy, as demonstrated in Table \ref{Table Transportation}. Slower modalities such as pedestrian movement (5 km/h) yield diminished completion rates (61.11\%), whereas expedited modalities including public transit (20 km/h) and private vehicular transport (30 km/h) achieve completion rates exceeding 95\%. These findings indicate that users should prioritize transportation modality selection based on task-specific requirements and performance objectives. Second, departure day analysis reveals substantial performance variations between weekdays and weekends, as illustrated in Table \ref{Table Departure Day}. The system attains optimal completion rates exceeding 96\% during weekdays, while weekend performance exhibits considerable degradation (Saturday: 75.84\%; Sunday: 50.59\%) attributable to the closure of numerous POIs during rest periods. Finally, temporal sensitivity analysis presented in Table \ref{Table Departure Time} demonstrates optimal system performance during mid-morning intervals (10:00-12:00) with completion rates surpassing 95\%, while early morning (08:00) and evening (18:00) departures manifest reduced performance metrics (75.44\% and 43.11\%, respectively). Our \Name system enables manual departure time modification and provides intelligent recommendations upon detecting POI closures, maintaining consistency with Google Maps functionality. Adaptive departure time optimization presents challenges, requiring balance between departure timing and dwell duration.

\subsection{Impact of User Location and POI Types}

\begin{figure}[h]
    \centering
    \subfigure[Shopping Mall - Rating]{\includegraphics[width=0.3\textwidth]{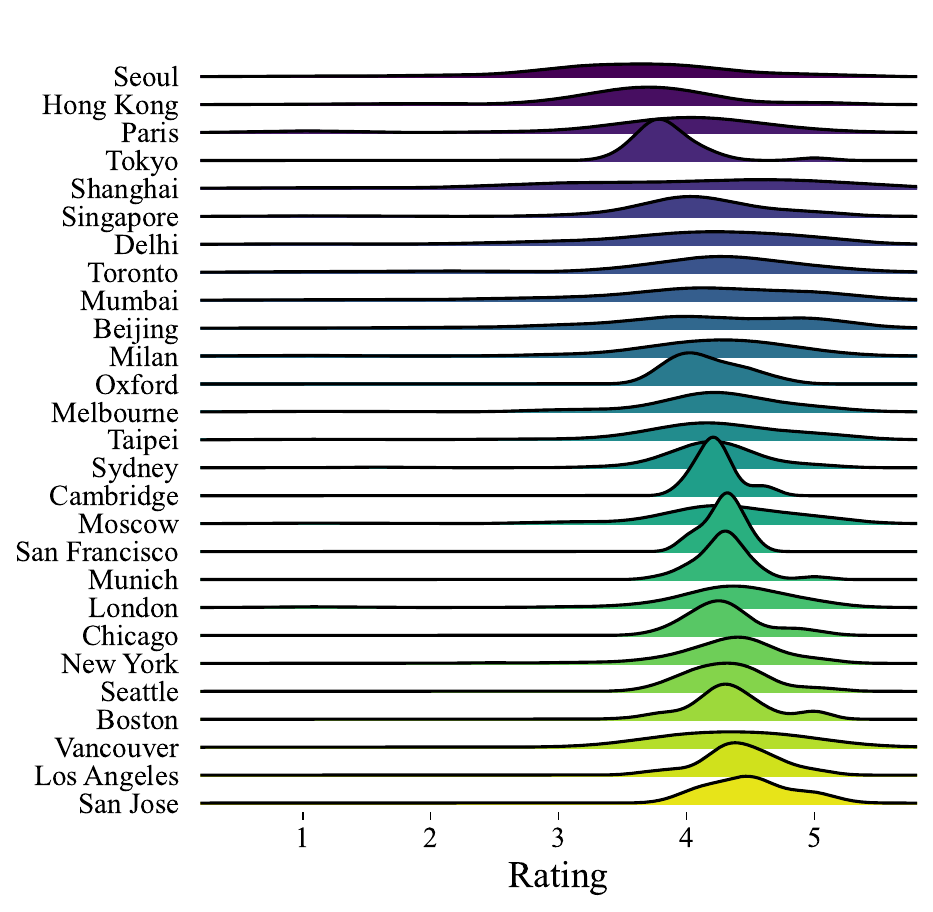}} \hspace{0.03\textwidth}
    \subfigure[Supermarket - Rating]{\includegraphics[width=0.3\textwidth]{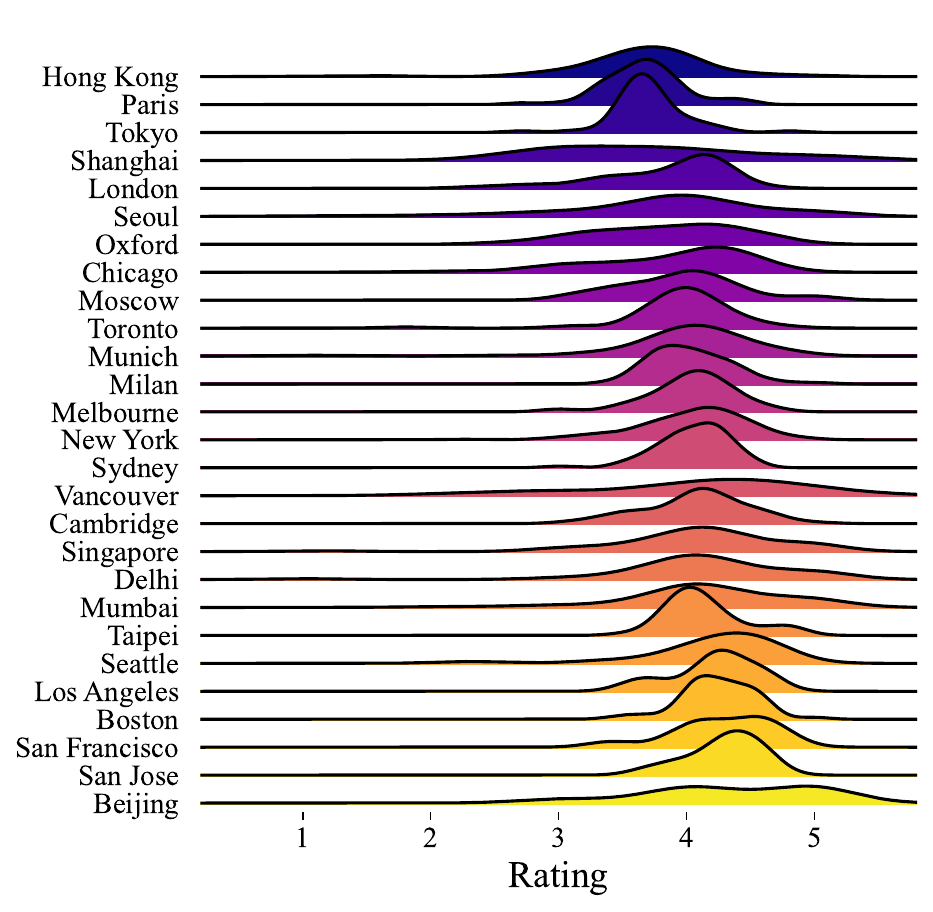}} \hspace{0.03\textwidth}
    \subfigure[Pharmacy - Rating]{\includegraphics[width=0.3\textwidth]{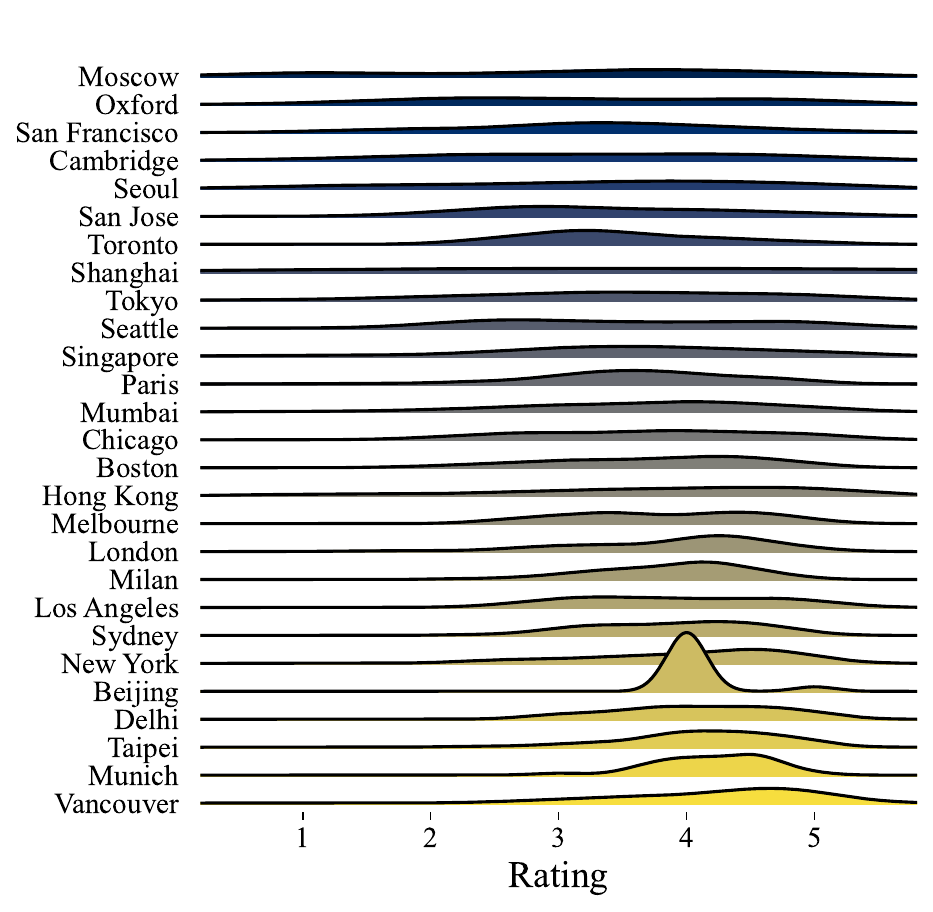}}

    \subfigure[\scriptsize Shopping Mall - Number of Reviews]{\includegraphics[width=0.3\textwidth]{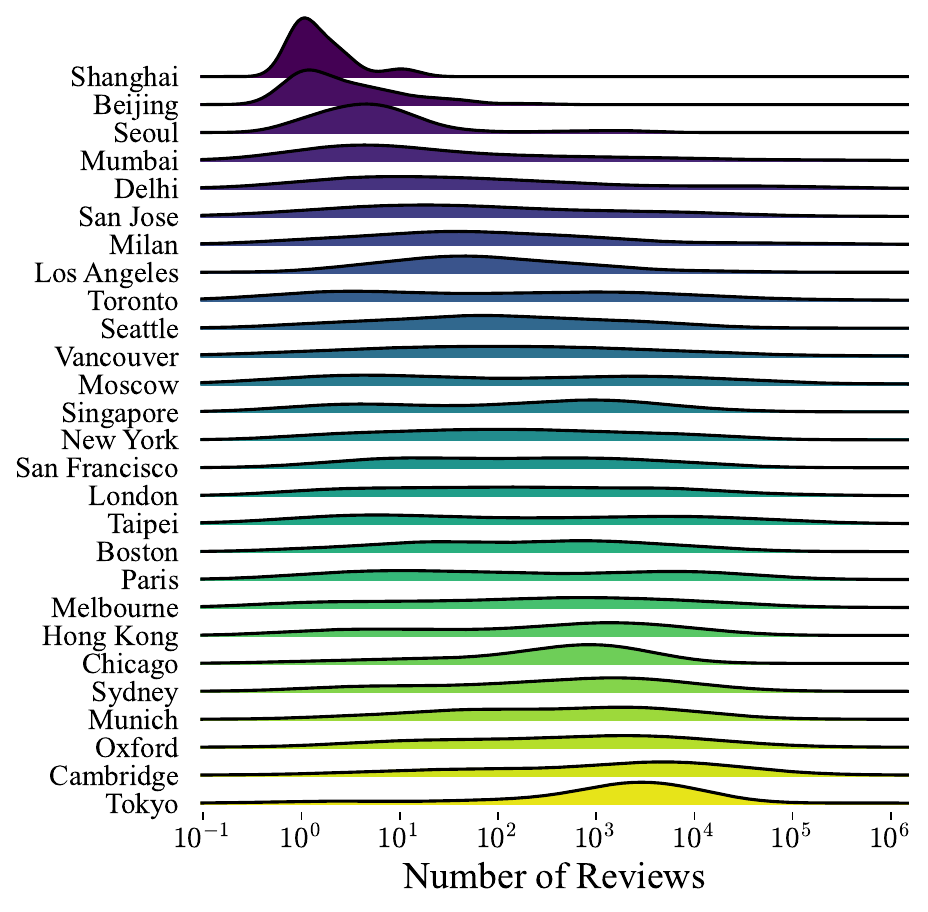}} \hspace{0.03\textwidth}
    \subfigure[Supermarket - Number of Reviews]{\includegraphics[width=0.3\textwidth]{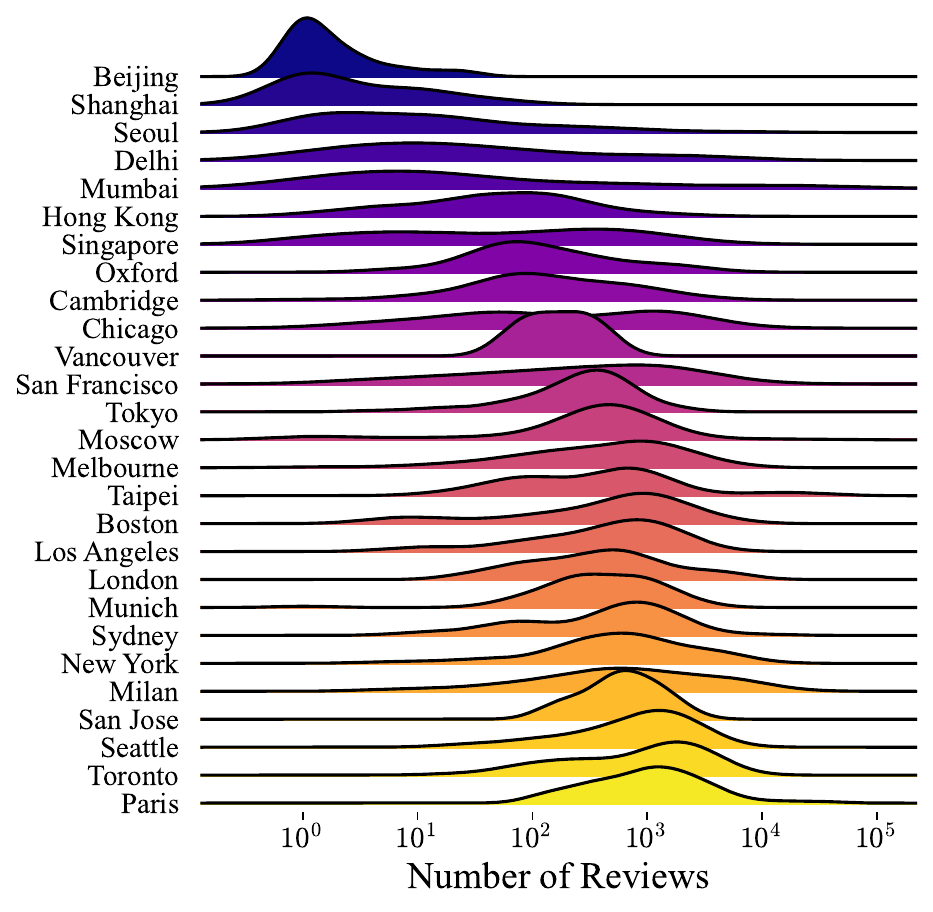}} \hspace{0.03\textwidth}
    \subfigure[Pharmacy - Number of Reviews]{\includegraphics[width=0.3\textwidth]{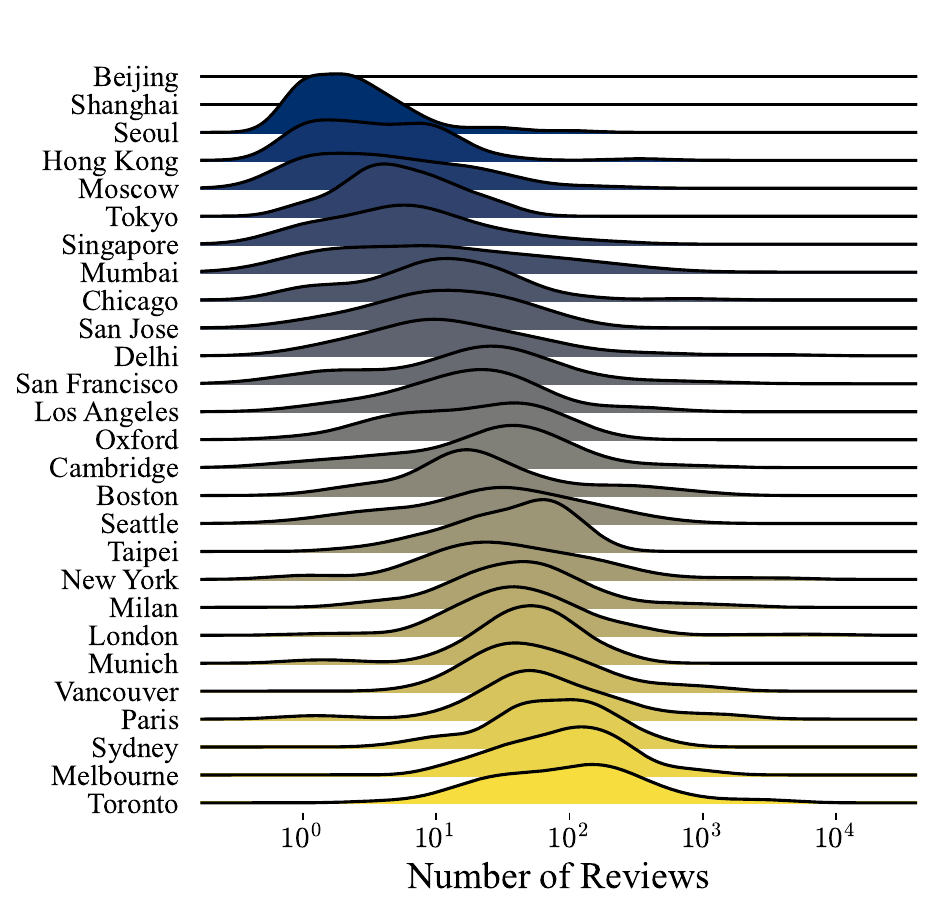}}

    \subfigure[Bank - Rating]{\includegraphics[width=0.3\textwidth]{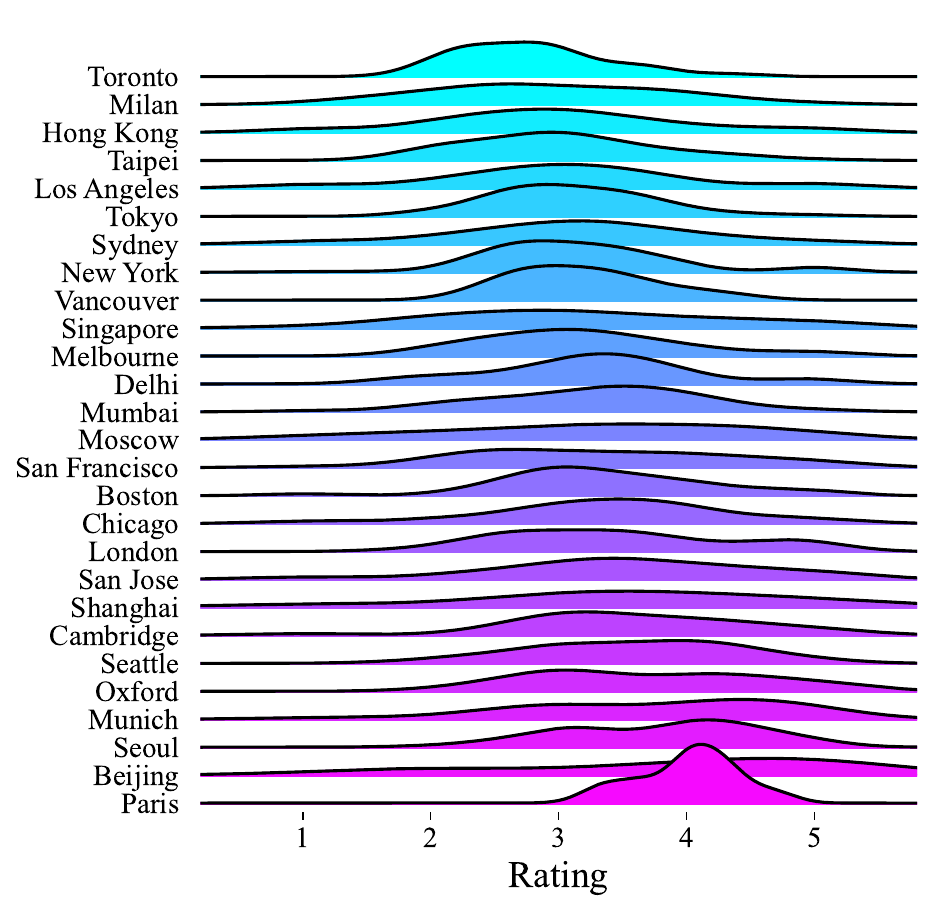}} \hspace{0.03\textwidth}
    \subfigure[Library - Rating]{\includegraphics[width=0.3\textwidth]{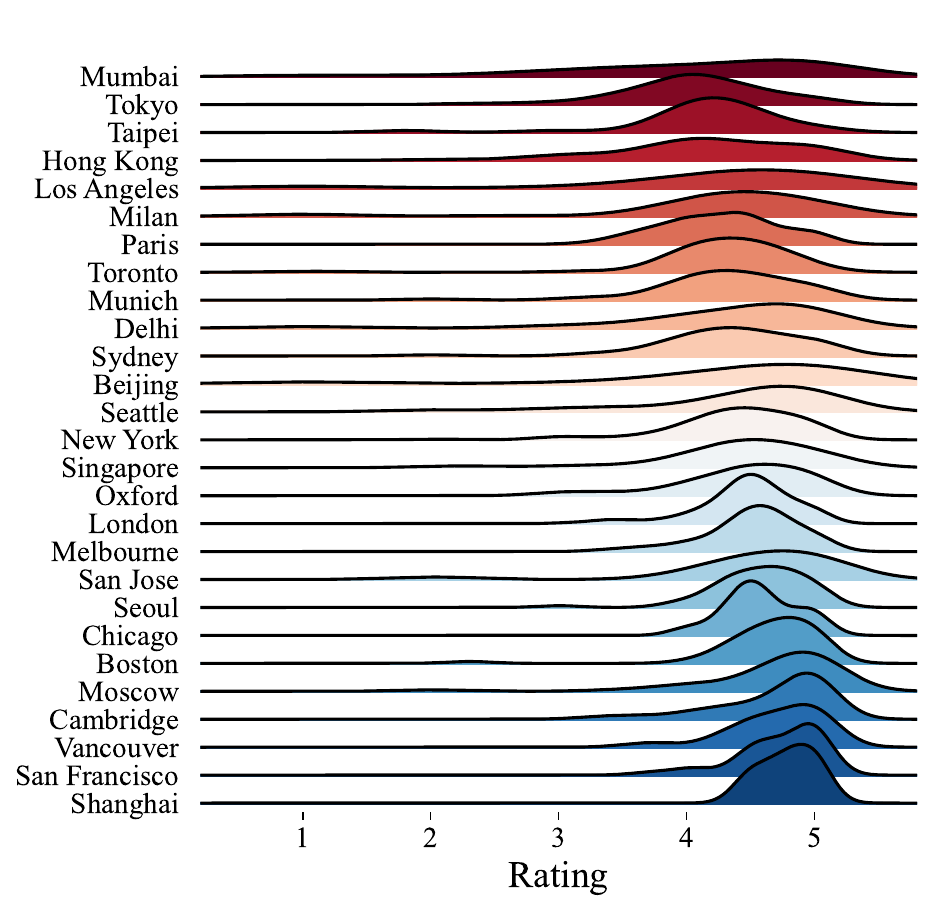}} \hfill \hfill
    
    \subfigure[Bank - Number of Reviews]{\includegraphics[width=0.3\textwidth]{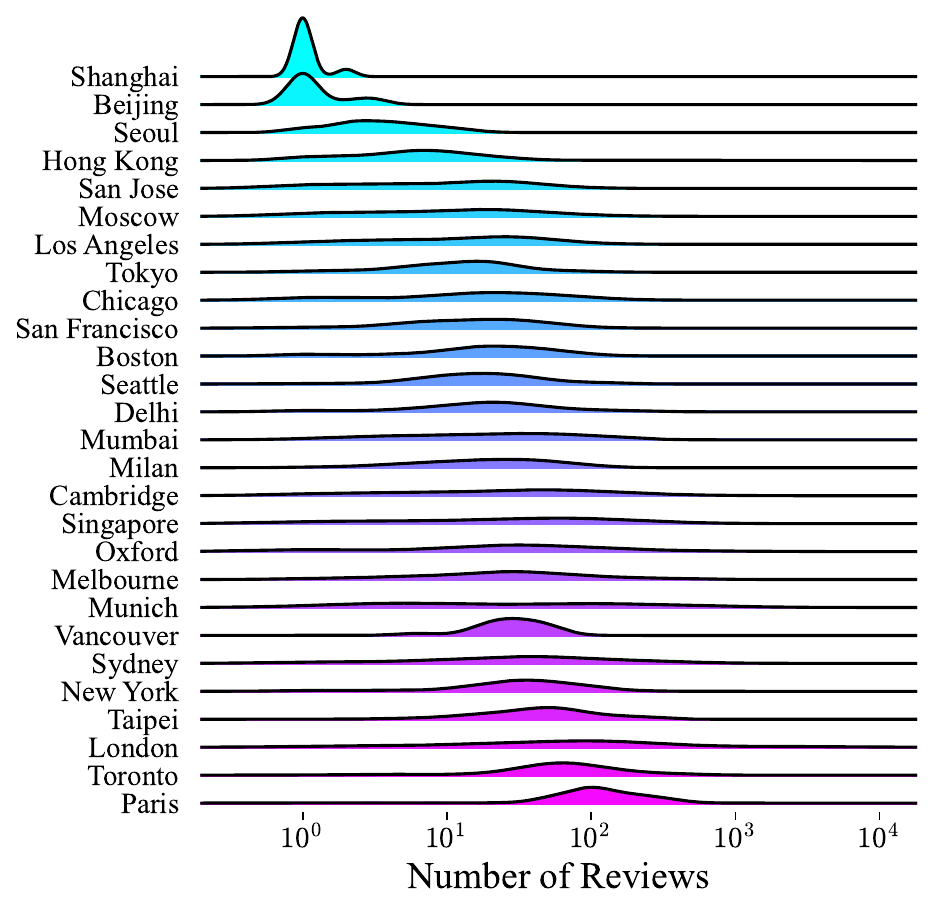}} \hspace{0.03\textwidth}
    \subfigure[Library - Number of Reviews]{\includegraphics[width=0.3\textwidth]{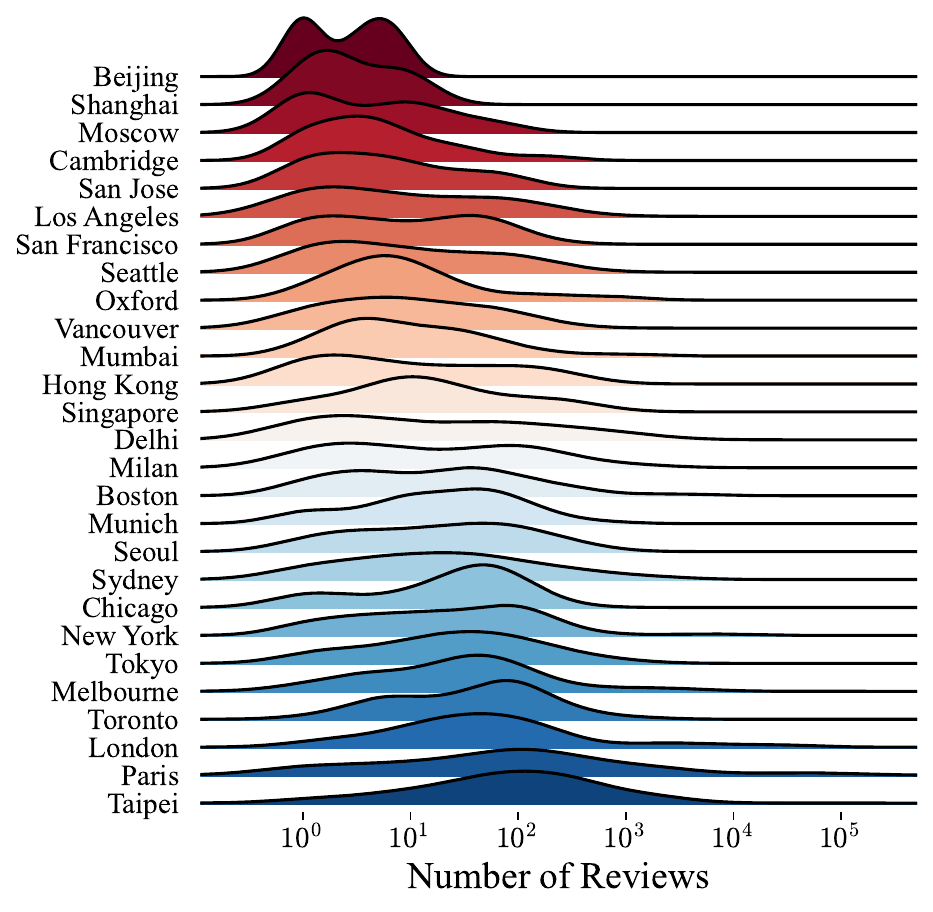}} \hfill \hfill
    \caption{Distribution of ratings and number of ratings for different POI types across different cities.}
    \label{fig:map_data}
\end{figure}

Figure \ref{fig:map_data} illustrates the distribution of POI ratings and review counts across different cities. The data reveals significant heterogeneity across cities and POI types, which presents a fundamental challenge in serving users worldwide. The frequency of Google Maps usage varies substantially across cities. For example, in Chinese metropolises like Beijing and Shanghai, despite their high population density, the number of reviews is notably lower due to Google Maps' limited popularity among Chinese users. However, international tourists visiting China may be unfamiliar with or have restricted access to local mapping services. A key advantage of our system is its ability to perform route planning on such unpredictable data distributions without requiring any training. Across different POI types, shopping malls and supermarkets exhibit review counts approximately an order of magnitude higher than pharmacies, banks, and libraries, primarily due to their significantly higher foot traffic. Benefit by the unique multi-step optimization design of our \AlgName algorithm, which prioritizes task completion rate before optimizing other metrics, these order-of-magnitude differences in POI types do not impede our ability to make appropriate POI selections.

\subsection{Integration of Essential Waypoints}

Our \Name system currently focuses on effectively removing POIs based on various constraints, while also possessing the potential to implement integration of essential waypoints functionality into route planning. This functionality requires domain-specific knowledge to formulate constraint rules for implementation. Similar to the three constraints established in the current \Name system, adding waypoints functionality can be categorized as originating from either user instructions or environmental information. For example, the vehicle's fuel level constitutes one of the constraints within the aforementioned environmental information. To develop an automatic gas station addition feature, specifically a fuel constraint, we need to determine that the current fuel level proves insufficient to complete the route based on vehicle fuel capacity, route length, and fuel consumption rate, thereby identifying the necessity to incorporate a gas station. Although intuitively one might anticipate that removing POIs and adding essential waypoints could generate conflicts, as they represent two distinct approaches to satisfying constraint conditions, our designed multi-step optimization strategy readily resolves these potential conflicts. Since, after fulfilling all constraint conditions, the subsequent step maximizes $\mathcal{O}_{\mathrm{(i)}}$ in Equation (\ref{eq:optimization}), which represents the task completion rate. Evidently, these newly incorporated essential waypoints, such as gas stations, do not constitute components of the task completion rate but have already been addressed within the constraints. Consequently, this strategic design establishes a sequential order for removing POIs and adding waypoints, ensuring that we do not eliminate task-relevant POIs to satisfy constraints requiring essential waypoint addition.

\onecolumn

\onecolumn

\section{Prompt Template}

\subsection{Human Instruction Generation}

\begin{tcolorbox}[boxrule = 0.5pt, breakable, box align=center,]
\textbf{Synthetic Label $\rightarrow$ Human Instruction (default with CoT prompting):}
\begin{lstlisting}[language=system prompt,escapeinside={(*}{*)}]
(*\textcolor{key}{"system\_prompt"}*) = 
```
You need to generate a natural human instruction by following these thinking steps:

1. First sentence: State all the POIs that need to be visited today:
   - Look at the list of POIs
   - Think about natural way to express visiting multiple POIs
   
2. Second sentence: State the return time if specified:
   - If specific time given, express as deadline
   - If no time limit, omit this part
   
3. Third sentence: Express POI rating vs route length preference indirectly:
   - If POI rating weight > 0.5: Express strong desire for high rating POIs
   - If route length weight > 0.5: Express urgency or need for efficiency
   - If balanced: Express desire for both reasonable rating and efficiency
   
4. Forth sentence: Express each dependency separately, one by one:
   - Convert each [A,B] pair into natural sequence requirement
   - Think about natural ways to express "must visit A before B"

Keep the language natural but always follow this structure. 
```
\end{lstlisting}

\begin{lstlisting}[language=user prompt,escapeinside={(*}{*)}]
(*\textcolor{key}{"user\_prompt"}*) =
```
Based on the following information, generate a natural human instruction:

1. POIs to visit: {', '.join(synthetic_label['pois'])}

2. Return by: {synthetic_label['time_limit']}

3. Preference analysis:
- POI rating weight: {synthetic_label['quality_weight']}
- Route length weight: {synthetic_label['distance_weight']}

4. Dependencies to express: {synthetic_label['dependencies']}

Generate ONE natural instruction that includes all this information. No prefix, additional text, or explanation.
```
\end{lstlisting}
\end{tcolorbox}
\vspace{-10pt}
\captionof{figure}{Prompt template for generating natural human instruction from synthetic label using GPT-4o with CoT prompting.}
\label{fig:prompt_for_instruction}

\subsection{LLM-as-Parser Estimation}

\begin{tcolorbox}[boxrule = 0.5pt, breakable, box align=center,]
\textbf{LLM-as-Parser -- Human Instruction $\rightarrow$ Estimation (without CoT prompting):}
\begin{lstlisting}[language=system prompt,escapeinside={(*}{*)}]
(*\textcolor{key}{"system\_prompt"}*) = 
```
Extract POIs, constraints, and weights from a human instruction. POIs should be a simple string array. Dependencies should be an empty list if no sequence requirements are mentioned. Weights should be between 0 and 1, and sum to 1. Extract them from language about importance of POI rating vs route length. 

Output must be valid JSON with structure:
{
    "pois": ["poi1", "poi2", ...],
    "time_limit": "HH:00" or "None",
    "dependencies": [["poi1", "poi2"], ...],
    "quality_weight": float,
    "distance_weight": float
}
```
\end{lstlisting}

\begin{lstlisting}[language=user prompt,escapeinside={(*}{*)}]
(*\textcolor{key}{"user\_prompt"}*) =
```
Extract POIs, constraints, and weights from this instruction as JSON:
{instruction}

Only output the JSON object. No prefix, additional text, or explanation.
```
\end{lstlisting}
\end{tcolorbox}
\vspace{-10pt}
\captionof{figure}{Prompt template for POI extraction, constraint identification, and user preference estimation using LLM-as-Parser without CoT prompting.}
\label{fig:parser_estimation}
\vspace{10pt}

\begin{tcolorbox}[boxrule = 0.5pt, breakable, box align=center,]
\textbf{LLM-as-Parser -- Human Instruction $\rightarrow$ Estimation (with CoT prompting):}
\begin{lstlisting}[language=system prompt,escapeinside={(*}{*)}]
(*\textcolor{key}{"system\_prompt"}*) = 
```
Extract POIs, constraints, and weights from human instruction through step-by-step reasoning:

1. POIs Analysis:
    - Look for POIs mentioned that need to be visited
    - Create list of unique POIs

2. Time Limit Analysis:
    - Search for any specific return time
    - Format as HH:00 or "None" if not specified

3. Dependencies Analysis:
    - Look for words indicating sequence (before, after, then, etc.)
    - Create pairs of [POI1, POI2] for each sequence requirement

4. Preference Analysis:
    - Look for language about POI quality/rating importance vs route efficiency
    - High POI rating emphasis (quality, best places, etc.) -> quality_weight should be large than 0.5
    - High route efficiency emphasis (quick, shortest, save time) -> distance_weight should be large than 0.5
    - Balanced language -> both weights around 0.5

Output must be valid JSON with structure:
{
    "pois": ["poi1", "poi2", ...],
    "time_limit": "HH:00" or "None",
    "dependencies": [["poi1", "poi2"], ...],
    "quality_weight": float,
    "distance_weight": float
}
```
\end{lstlisting}

\begin{lstlisting}[language=user prompt,escapeinside={(*}{*)}]
(*\textcolor{key}{"user\_prompt"}*) =
```
{instruction}

Only output the list object. No prefix, additional text, or explanation.
```
\end{lstlisting}
\end{tcolorbox}
\vspace{-10pt}
\captionof{figure}{Prompt template for POI extraction, constraint identification, and user preference estimation using LLM-as-Parser with CoT prompting.}
\label{fig:parser_estimation_CoT}

\subsection{LLM-as-Agent Estimation}

\begin{tcolorbox}[boxrule = 0.5pt, breakable, box align=center,]
\textbf{LLM-as-Agent -- Human Instruction $\rightarrow$ Estimation (without CoT prompting):}
\begin{lstlisting}[language=system prompt,escapeinside={(*}{*)}]
(*\textcolor{key}{"system\_prompt"}*) = 
```
You are a route planning assistant. Your goal is to plan an optimal route based on the following objectives:

Primary Objectives:
1. Minimize the total route length/distance 
2. Maximize coverage of different POI types (select exactly one POI per required type)
3. Maximize the quality of visited POIs (based on ratings and number of ratings)
4. Balance between route efficiency and POI quality
5. Ensure compliance with time limits from instructions
6. Account for dependencies between POIs
7. Respect opening hours of recommended POIs

Visit Duration: shopping mall: 120 mins, supermarket: 30 mins, pharmacy: 15 mins, bank: 20 mins, library: 60 mins
Travel Speed: 30 km/h
Departure Time: 10:00 AM

Always output POI IDs as provided in the input data. Your output must strictly follow this format:
[POI ID, POI ID, POI ID]

Available POIs:
    POI ID: ChIJQ0eRlQ1644kR11stZbdBvM0:
    * Type: shopping mall
    * Rating: 4.4 (4110 reviews)
    * Coordinates: (42.3471832, -71.0778024)
    * Opening Hours: Monday: 11:00 AM - 7:00 PM

    (*\textcolor{RoyalBlue}{...... \# Omitting the remaining 9 shopping malls}*)

    POI ID: ChIJ296XtDV344kRXTFzOH4n2Vg:
    * Type: supermarket
    * Rating: 4.4 (4547 reviews)
    * Coordinates: (42.38078309999999, -71.10163109999999)
    * Opening Hours: Monday: 7:00 AM - 9:00 PM

    (*\textcolor{RoyalBlue}{...... \# Omitting the remaining 9 supermarkets}*)
    
    POI ID: ChIJf52pw8lw44kRzVIBthOOKgM:
    * Type: pharmacy
    * Rating: 4.1 (38 reviews)
    * Coordinates: (42.37640409999999, -71.0901677)
    * Opening Hours: Monday: 9:00 AM - 1:30 PM, 2:00 - 7:00 PM

    (*\textcolor{RoyalBlue}{...... \# Omitting the remaining 9 pharmacies}*)

    POI ID: ChIJXcQ5xYNw44kRvDwvuZw29wI:
    * Type: bank
    * Rating: 4.5 (114 reviews)
    * Coordinates: (42.35561349999999, -71.0612641)
    * Opening Hours: Monday: 9:00 AM - 5:00 PM

    (*\textcolor{RoyalBlue}{...... \# Omitting the remaining 9 banks}*)

    POI ID: ChIJU_ibQgx644kRc6zBW_5d1Lk:
    * Type: library
    * Rating: 4.8 (2770 reviews)
    * Coordinates: (42.3493136, -71.0781875)
    * Opening Hours: Monday: 9:00 AM - 8:00 PM

    (*\textcolor{RoyalBlue}{...... \# Omitting the remaining 9 libraries}*)
```
\end{lstlisting}

\begin{lstlisting}[language=user prompt,escapeinside={(*}{*)}]
(*\textcolor{key}{"user\_prompt"}*) =
```
{instruction}

Only output the list object. No prefix, additional text, or explanation.
```
\end{lstlisting}
\end{tcolorbox}
\vspace{-10pt}
\captionof{figure}{Prompt template for POI extraction, constraint identification, and user preference estimation using LLM-as-Agent without CoT prompting. Due to input token length limitations, we sample 10 specific POIs for each POI type as input for LLM-as-Agent.}
\label{fig:agent_estimation}
\vspace{10pt}

\begin{tcolorbox}[boxrule = 0.5pt, breakable, box align=center,]
\textbf{LLM-as-Agent -- Human Instruction $\rightarrow$ Estimation (with CoT prompting):}
\begin{lstlisting}[language=system prompt,escapeinside={(*}{*)}]
(*\textcolor{key}{"system\_prompt"}*) = 
```
You are a route planning assistant. Your goal is to plan an optimal route based on the following objectives:

Primary Objectives:
1. Minimize the total route length/distance 
2. Maximize coverage of different POI types (select exactly one POI per required type)
3. Maximize the quality of visited POIs (based on ratings and number of ratings)
4. Balance between route efficiency and POI quality
5. Ensure compliance with time limits from instructions
6. Account for dependencies between POIs
7. Respect opening hours of recommended POIs

Visit Duration: shopping mall: 120 mins, supermarket: 30 mins, pharmacy: 15 mins, bank: 20 mins, library: 60 mins
Travel Speed: 30 km/h
Departure Time: 10:00 AM

Planning Process:
1. Analyze User Requirements:
  - Identify required POI types from human instruction
  - Note any specified preferences for particular types

2. Prioritize POI Types:
  - Order POI types based on:
    * User specified preferences/requirements
    * Dependencies between types
    * Opening hours

3. Select Specific POIs:
  - For each POI type in order:
    * Consider only POIs of that specific type
    * Choose exactly one POI based on:
      - Rating and number of reviews
      - Location efficiency (distance to previous/next points)
      - Opening hours compatibility
  - Ensure only one POI is selected per type

First analyze the constraints and requirements, then plan accordingly. After planning, validate that your route satisfies all constraints.

Always output POI IDs as provided in the input data. Your output must strictly follow this format:
[POI ID, POI ID, POI ID]

Available POIs:
    POI ID: ChIJQ0eRlQ1644kR11stZbdBvM0:
    * Type: shopping mall
    * Rating: 4.4 (4110 reviews)
    * Coordinates: (42.3471832, -71.0778024)
    * Opening Hours: Monday: 11:00 AM - 7:00 PM

    (*\textcolor{RoyalBlue}{...... \# Omitting the remaining 9 shopping malls}*)

    POI ID: ChIJ296XtDV344kRXTFzOH4n2Vg:
    * Type: supermarket
    * Rating: 4.4 (4547 reviews)
    * Coordinates: (42.38078309999999, -71.10163109999999)
    * Opening Hours: Monday: 7:00 AM - 9:00 PM

    (*\textcolor{RoyalBlue}{...... \# Omitting the remaining 9 supermarkets}*)
    
    POI ID: ChIJf52pw8lw44kRzVIBthOOKgM:
    * Type: pharmacy
    * Rating: 4.1 (38 reviews)
    * Coordinates: (42.37640409999999, -71.0901677)
    * Opening Hours: Monday: 9:00 AM - 1:30 PM, 2:00 - 7:00 PM

    (*\textcolor{RoyalBlue}{...... \# Omitting the remaining 9 pharmacies}*)

    POI ID: ChIJXcQ5xYNw44kRvDwvuZw29wI:
    * Type: bank
    * Rating: 4.5 (114 reviews)
    * Coordinates: (42.35561349999999, -71.0612641)
    * Opening Hours: Monday: 9:00 AM - 5:00 PM

    (*\textcolor{RoyalBlue}{...... \# Omitting the remaining 9 banks}*)

    POI ID: ChIJU_ibQgx644kRc6zBW_5d1Lk:
    * Type: library
    * Rating: 4.8 (2770 reviews)
    * Coordinates: (42.3493136, -71.0781875)
    * Opening Hours: Monday: 9:00 AM - 8:00 PM

    (*\textcolor{RoyalBlue}{...... \# Omitting the remaining 9 libraries}*)
```
\end{lstlisting}

\begin{lstlisting}[language=user prompt,escapeinside={(*}{*)}]
(*\textcolor{key}{"user\_prompt"}*) =
```
{instruction}

Only output the list object. No prefix, additional text, or explanation.
```
\end{lstlisting}
\end{tcolorbox}
\vspace{-10pt}
\captionof{figure}{Prompt template for POI extraction, constraint identification, and user preference estimation using LLM-as-Agent with CoT prompting. Due to input token length limitations, we sample 10 specific POIs for each POI type as input for LLM-as-Agent.}
\label{fig:agent_estimation_CoT}

\end{document}